\documentclass[10pt,twocolumn,letterpaper]{article}

\usepackage[pagenumbers]{iccv} 

\definecolor{iccvblue}{rgb}{0.21,0.49,0.74}
\usepackage[pagebackref,breaklinks,colorlinks,allcolors=iccvblue]{hyperref}
\usepackage{multirow}
\usepackage{multicol}
\definecolor{mygrey}{RGB}{239,239,239}
\usepackage{colortbl}
\usepackage[accsupp]{axessibility}  

\def\paperID{xxxx} 
\def\confName{ICCV}
\def\confYear{2025}

\title{GeoExplorer: Active Geo-localization with Curiosity-Driven Exploration}

\author{Li Mi, Manon Béchaz, Zeming Chen, Antoine Bosselut, Devis Tuia\\
\quad
EPFL, Switzerland\\
{\tt\small \url{https://limirs.github.io/GeoExplorer/}}
}

\begin{document}
\maketitle

\begin{abstract}
Active Geo-localization (AGL) is the task of localizing a goal, represented in various modalities (e.g., aerial images, ground-level images, or text), within a predefined search area. Current methods approach AGL as a goal-reaching reinforcement learning (RL) problem with a distance-based reward. They localize the goal by implicitly learning to minimize the relative distance from it. However, when distance estimation becomes challenging or when encountering unseen targets and environments, the agent exhibits reduced robustness and generalization ability due to the less reliable exploration strategy learned during training. In this paper, we propose GeoExplorer, an AGL agent that incorporates curiosity-driven exploration through intrinsic rewards. Unlike distance-based rewards, our curiosity-driven reward is goal-agnostic, enabling robust, diverse, and contextually relevant exploration based on effective environment modeling. These capabilities have been proven through extensive experiments across four AGL benchmarks, demonstrating the effectiveness and generalization ability of GeoExplorer in diverse settings, particularly in localizing unfamiliar targets and environments.
\end{abstract}

\section{Introduction}
\label{sec:intro}
Guiding unmanned aerial vehicles (UAVs) is essential in search-and-rescue operations, which necessitate UAVs to screen affected areas and locate missing individuals or damaged structures~\cite{meera2019obstacle,narazaki2022vision}. Instead of conducting exhaustive searches across entire regions, UAVs can be actively guided using available information: one could use coordinates of the point of interest, historical aerial imagery, or even ground photographs or textual descriptions~\cite{sarkar2024gomaa, pirinen2022aerial}.

\begin{figure}[t]
    \centering
    \includegraphics[width=0.99\linewidth]{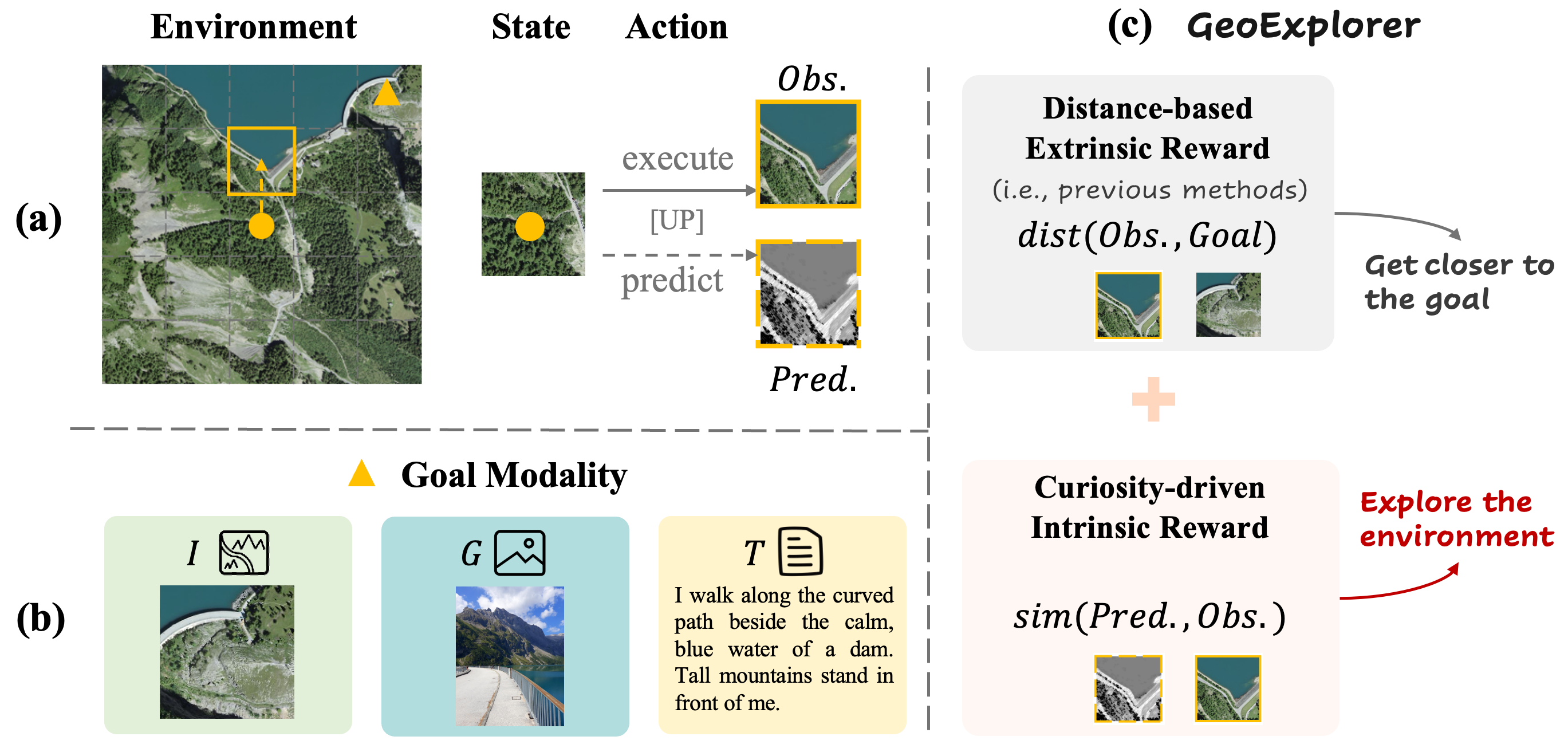}
    \caption{\textbf{GeoExplorer combines goal-oriented and curiosity-driven rewards to address AGL.} (a) \textit{Task setting.} Given a starting patch ($\circ$) and a predefined search area (\emph{environment}), AGL aims at localizing a goal ($\triangle$), by guiding an agent towards it. At a given time, the agent observes a \emph{state} and selects an \emph{action}. The goal location is \textit{unknown} for inference. (b) \textit{Goal modalities.} The goal can be presented in various modalities including aerial images (I), ground-level images (G) and text (T), associated with a patch in the area. (c) \textit{GeoExplorer.} The learning process of GeoExplorer is guided by two rewards: \textbf{1)} a goal-oriented \textbf{dist}ance-based reward, as in previous methods, which directs the agent toward the goal, and \textbf{2)} a goal-agnostic, curiosity-driven reward that encourages the agent to explore the environment based on \textit{curiosity} measured by \textbf{sim}ilarity of state prediction and observation.} 
    \label{fig:intro}
\end{figure}

The problem is known in computer vision as Active Geo-Localization (AGL)~\cite{sarkar2024gomaa, pirinen2022aerial} within a goal-reaching reinforcement learning (RL) context (Figure~\ref{fig:intro}~(a)). AGL focuses on localizing a target (\emph{goal}), within a predefined search area (\emph{environment}) presented in the bird's eye view, by navigating the agent towards it. At a given time, the agent observes a \emph{state}, \textit{i.e.}, a patch representing a limited observation of the environment, and selects an \emph{action}, \textit{i.e.}, a decision that modifies the agent position and the observed state.
The goal content can be described in various modalities (\textit{e.g.}, aerial patches, ground-level images, and textual descriptions, Figure~\ref{fig:intro}~(b)), but its location is \textit{unknown} during inference. The task is considered successful if the agent reaches the goal location within a limited search budget ($\mathcal{B}$). In real-world applications of AGL, robustness and generalizability are crucial as few scenarios are identical~\cite{sarkar2024gomaa}.

Existing AGL methods rely on extrinsic rewards (\textit{i.e.}, goal-oriented), and the policy is implicitly shaped by minimizing the relative distance between the agent and the goal during training. However, this distance is \textit{unknown} during inference (since the goal's location is unknown), making the learned strategy less reliable. Moreover, these distance estimations are often tailored to specific training environments, making it challenging for models to adapt to unseen goals or new environments.  
Curiosity-driven intrinsic rewards (\textit{i.e.}, goal-agnostic) offer a compelling solution. 
By rewarding the agent for encountering less familiar~\cite{pathak2017curiosity, burda2018large} or unpredictable~\cite{stadie2015incentivizing, pathak2017curiosity} state transitions, the intrinsic reward encourages \textit{exploration of the environment}~\cite{pathak2017curiosity, kim2020active}. 
Therefore, integrating curiosity-driven rewards with goal-oriented rewards introduces an essential trade-off between following the direct, goal-oriented guidance and engaging in exploratory behavior. This balance ensures that the agent not only efficiently reaches its goals but also continually refines its understanding of the environment, leading to effective and adaptive policies which enhance \textit{robustness and generalization} in AGL.

\begin{figure}[t]
    \centering
    \includegraphics[width=0.98\linewidth]{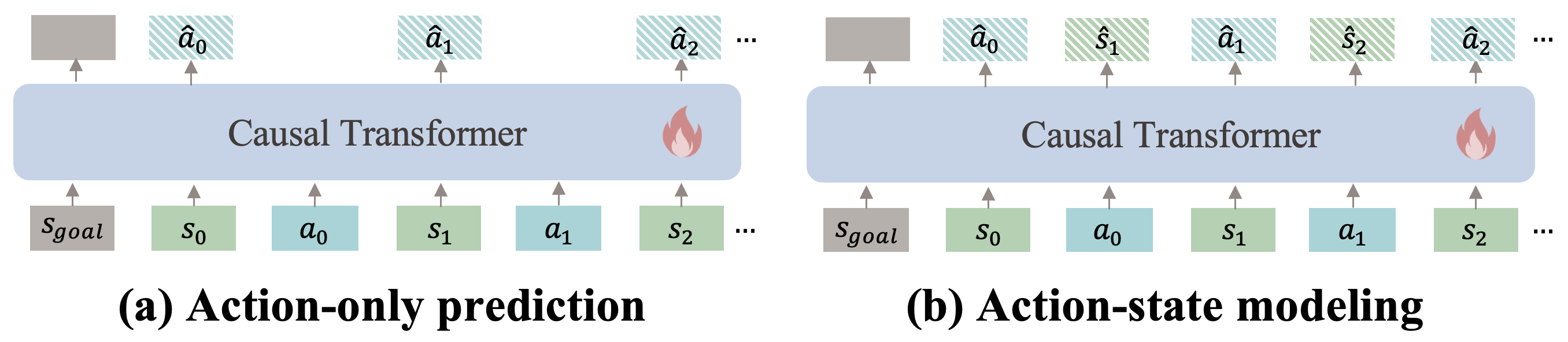}
    \caption{Comparison between (a) action-only prediction and (b) joint action-state modeling.}
    \label{fig:action_state_modeling}
\end{figure}

However, constructing intrinsic rewards requires a way to \textit{predict} state transitions, as it is based on the discrepancy between state prediction and actual observation. Previous AGL methods~\cite{pirinen2022aerial, sarkar2024gomaa} only focus on predicting the action sequence that leads to the goal~\cite{chen2021decision}  (Figure~\ref{fig:action_state_modeling} (a)), and therefore do not satisfy this requirement. Without explicit state prediction, these methods also fail to anticipate how an action can change the state accurately, resulting in reduced adaptability and accumulated errors in unseen environments.
Instead, joint action-state modeling (Figure~\ref{fig:action_state_modeling} (b)) encourages a more robust and adaptive environment modeling and representation refinement~\cite{wang2023mimicplay, bar2024navigation}, which is crucial when facing unfamiliar environments or goals.

In this work, we introduce GeoExplorer, an AGL agent that \textbf{1)} jointly predicts the \textit{action-state dynamics} and \textbf{2)} explores the search region with a \textit{curiosity-driven reward}.
The learning process of GeoExplorer boils down to two stages: \textit{Action-State Dynamics Modeling (DM)} and \textit{Curiosity-Driven Exploration (CE)}. 
DM simultaneously models action-state dynamics using a causal Transformer and allows GeoExplorer not only to learn which actions lead to the goal, but also how these actions influence state transitions, thus providing a richer, more comprehensive representation of the environment. 
Then, during CE, an intrinsic reward is introduced to guide the exploration process by leveraging the discrepancy between the state prediction provided by DM and the real-world state observation. 
Experimental results on four datasets, including a \textit{novel unseen target generalization} setting, demonstrate that GeoExplorer provides dense, goal-agnostic and content-aware rewards for AGL. 
Combined with goal-oriented guidance, GeoExplorer highlights that balancing \textit{goal-reaching} (extrinsic reward) and \textit{environment exploration} (intrinsic reward) can improve both localization performance and generalization ability of the AGL agent. 
The resulting active and adaptive exploration ability of GeoExplorer represents a step towards deploying AGL in real-world search-and-rescue operations.

\section{Related Work}
\label{sec:related_work}

\noindent \textbf{Active Geo-Localization.} 
Although related, AGL differs from traditional visual geo-localization~\cite{zamir2016introduction} (\textit{i.e.}, typically focuses on identifying the location of a target through coordinate prediction~\cite{vivanco2023geoclip, astruc2024openstreetview} or retrieval~\cite{berton2022rethinking, berton2022deep, workman2015wide, zhu2021vigor, wang2023fine, mi2024congeo, xia2025fg}), by actively navigating the agent to reach the target. Current methods in AGL have made impressive progress in learning the action transitions from various UAV localization trajectories~\cite{pirinen2022aerial, sarkar2024gomaa}. Pirinen \textit{et al.}~\cite{fan2022aerial} framed the AGL as a goal-reaching RL problem, guided by a sparse extrinsic reward. Sarkar \textit{et al.}~\cite{sarkar2024gomaa} extended the AGL task with multiple goal modalities and proposed goal-aware action modeling pretraining and an actor-critic RL pipeline with distance-based extrinsic reward. Despite progress, these methods localize the goal guided by extrinsic reward, which might provide misleading guidance when the target's location is unknown. Moreover, this approach fails to effectively and comprehensively model the environment, leading to inefficient search strategies and generalization to unseen regions. GeoExplorer aims to address these challenges by modeling action-state dynamics simultaneously and enabling curiosity-driven exploration.

\begin{figure*}
    \centering
    \includegraphics[width=0.99\linewidth]{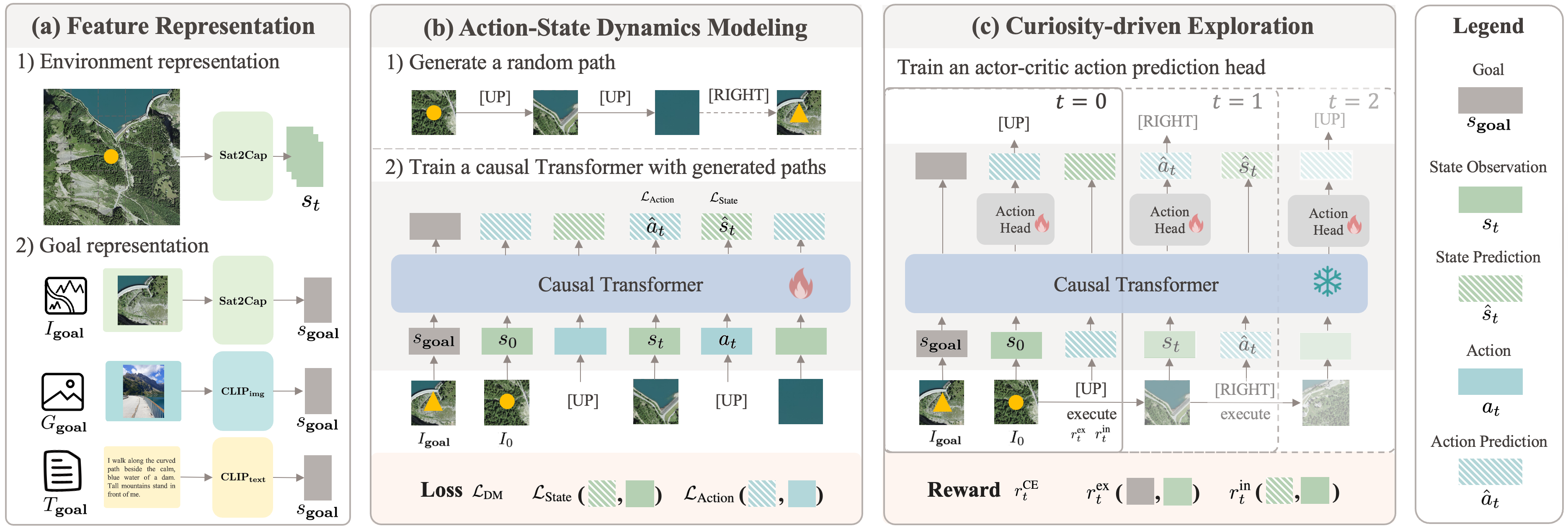}
    \caption{\textbf{GeoExplorer learning pipeline.} The learning process can be divided into three stages sequentially: Feature Representation, Action-State Dynamics Modeling (DM), and Curiosity-Driven Exploration (CE). (a) \textit{Feature Representation.} The environment ($s_t$) and goal ($s_\mathbf{goal}$) are encoded with different but aligned encoders, according to their modalities (\textit{e.g.}, aerial images ($I_\mathbf{goal}$), ground-level images ($G_\mathbf{goal}$), or text ($T_\mathbf{goal}$)). (b) \textit{Action-State Dynamics Modeling.} A causal Transformer is trained to jointly capture action-state dynamics, guided by supervision from generated action-state trajectories for environment modeling. (c) \textit{Curiosity-Driven Exploration.} Based on state prediction from (b), a curiosity-driven intrinsic reward ($r^{\mathbf{in}}_t$) is combined with the distance-based extrinsic reward ($r^{\mathbf{ex}}_t$) to encourage the agent to explore the environment by measuring the differences between prediction and observations.}
    \label{fig:pipeline}
\end{figure*}

\noindent \textbf{Transformers for Action-State Modeling.} Recent progress in RL has shown exciting potential in predicting actions through sequence modeling~\cite{chen2021decision, janner2021offline} based on causal Transformers~\cite{waswani2017attention, radford2019language}. They focus on predictive modeling of action sequences conditioned on a task specification (\textit{e.g.}, target goal or returns) as opposed to explicitly learning Q-functions or policy gradients~\cite{schmidhuber2019reinforcement,srivastava2019training, emmons2022rvs}.
Moreover, recent work has focused on learning \emph{world models} to describe environments. They further use transformer-style architecture to model the effect of the agent’s action on the environment, by predicting future state representations~\cite{wang2023mimicplay, bar2024navigation, zhou2024dino} or observations~\cite{yang2024learning, wu2024unleashing}. They often take the previous state history and a potential action as input and predict the next state. 
However, those world models are primarily designed for domains with \textit{continuous state transitions}, such as autonomous driving and robotics. In contrast, the AGL setting may involve non-overlapping and abrupt state changes by taking an action. GeoExplorer demonstrates that even in such cases, state modeling enhances action prediction, further underscoring the critical role of comprehensive environment modeling.

\noindent \textbf{Curiosity-Driven Reinforcement Learning.}
Curious exploration plays a critical role in RL by enabling agents to navigate and learn in environments where rewards are far in the future, sparse, or non-existent. Even when the extrinsic goal is well-defined, researchers have argued for the use of intrinsic motivation as an additional reward encouraging exploration~\cite{schmidhuber1991possibility}. Different approaches to generate intrinsic rewards have been explored, including encouraging exploration of previously unencountered states~\cite{pathak2017curiosity, burda2018large}, fostering curiosity when predictions about future states and the outcomes of actions differ~\cite{stadie2015incentivizing, pathak2017curiosity}, promoting exploration in areas of high uncertainty~\cite{kim2020active} and encouraging agents to align their behaviors with the expectations of others~\cite{ma2022elign}. 
Building upon these advances, we introduce curiosity-driven exploration to AGL, where adaptive and active exploration is essential for generalization. By rewarding the disagreement between predicted and actual states, GeoExplorer exemplifies the effective integration of curiosity-driven RL into sequence modeling frameworks, without additional components.

\section{GeoExplorer}
\label{sec:method}

\subsection{Overview}
\noindent \textbf{Problem Definition.} Following previous work~\cite{pirinen2022aerial, sarkar2024gomaa}, the AGL task can be considered as a goal-reaching RL problem.
A UAV agent is located at a starting patch ($I_{0}$) and aims at navigating within a predefined non-overlapping $X$ $\times$ $Y$ search grid to reach a goal location within a search budget ($\mathcal{B}$).
At each time step $t$, the agent observes only its current patch $I_t$ and can move between patches by taking actions $a_{t} \in \mathcal{A}$ (UP, DOWN, LEFT, RIGHT, when considering a bird's-eye-view). The state of time step $t$ can be denoted as $s_{t}$, which is a representation of the current patch. The overall objective of AGL is to determine a trajectory $x = \{s_{0}, a_{0}, \cdots, s_{t}, a_{t}, \cdots, s_\mathbf{goal}\}$ that leads the agent to the goal. During training, the goal is exclusively presented as an aerial view ($I_\mathbf{goal}$) with known location. For inference, the goal can be provided in one of the three modalities: aerial view, ground images ($G_\mathbf{goal}$) or text ($T_\mathbf{goal}$), but the location of the goal is unknown.

\noindent \textbf{Method Overview.} The training process of GeoExplorer can be divided into three stages (Figure~\ref{fig:pipeline}): 1) \textit{Feature Representation}, 2) \textit{Action-State Dynamics Modeling (DM)}, and 3) \textit{Curiosity-Driven Exploration (CE)}. DM is a supervised pretraining stage, where action-state dynamics are modeled by a causal Transformer. 
CE is a reinforcement learning stage, where an actor-critic network, used as an action prediction head, is trained to predict actions based on the output of the DM stage, combining both distance-based reward and curiosity-driven reward.

\subsection{Feature Representation}
In AGL, the goals are presented in different modalities. To achieve localization regardless of their nature, pre-aligned feature encoders are used for each modality (\textit{i.e.}, aerial images, ground-level images, and text), as in previous work~\cite{sarkar2024gomaa} and as depicted in Figure~\ref{fig:pipeline} (a). Those modality-specific encoders are pre-trained to align within a joint embedding space, with text-image pairs or image-image pairs through contrastive learning~\cite{radford2021learning, dhakal2024sat2cap}. More specifically, the ground image encoder (a Vision Transformer, $\mathbf{CLIP_{img}}$) and text encoder ($\mathbf{CLIP_{text}}$) are issued from the pre-trained CLIP model~\cite{radford2021learning}, while the aerial encoder ($\mathbf{Sat2Cap_{img}}$) is a ViT aligned with the CLIP image encoder using ground-level image and aerial image pairs~\cite{dhakal2024sat2cap}. The process can be denoted as:
\begin{align}
\label{eq:goal}
    s_\mathbf{goal} &= \mathbf{Sat2Cap_{img}}(I_\mathbf{goal}), \notag \\ 
    \text{or} \quad s_\mathbf{goal} &= \mathbf{CLIP_{img}}(G_\mathbf{goal}), \\
    \text{or} \quad s_\mathbf{goal} &= \mathbf{CLIP_{text}}(T_\mathbf{goal}). \notag
\end{align}
Moreover, we also encode the environment using the same aerial image encoder, and the agent's location at time step $t$, patch $I_{t}$, can be encoded as current state $s_{t}$ as: $s_{t} = \mathbf{Sat2Cap_{img}}(I_{t})$.

\subsection{Action-State Dynamics Modeling}
Different from previous efforts that use causal Transformers ($\mathrm{CausalTrans}$, \textit{e.g.}, LLMs) to predict \textit{actions only}~\cite{chen2021decision, sarkar2024gomaa}, we simultaneously model the \textit{action-state dynamics} without modifying the model architecture. 

\noindent \textbf{Random Path Generation.}
To collect action-state sequences for training, for each search area, we generate trajectories with random \{start, goal\} pairs, as in ~\cite{sarkar2024gomaa} (shown in Figure~\ref{fig:pipeline} (b)). Each generated sequence, denoted as $x_{n} = \{s_{0}, a_{0}, s_{1}, a_1, \ldots,s_t,a_t, \ldots,s_N, a_N\}$, consists of a series of randomly chosen actions $\{a_0, a_1,\ldots a_t,\ldots, a_{N}\}$ that explore possible transitions between states. Additionally, we define a ground truth action transitions $\{a^{*}_0, a^{*}_1,\ldots a^*_t,\ldots, a^{*}_{N}\}$ for each tracjectory. Each action $a^{*}_{t}$ represents an ideal goal-directed action which moves the agent from patch $I_{t}$ closer to the goal patch $I_\mathbf{goal}$. 

\noindent \textbf{Causal Transformer for Sequence Modeling.}
With the generated trajectories, we train a causal Transformer on a sequence modeling task that \textbf{1)} predicts the optimal actions $a^*_{t}$ that will bring the agent closer to the goal location; \textbf{2)} predicts a state representation $s_{t}$ at time $t$ that depends on the action at the previous time step $a_{t-1}$. The modeling process, illustrated in Figure~\ref{fig:pipeline} (b), can be denoted as:
\begin{align}
\hat{a}_t &= \mathrm{CausalTrans}(s_{t}|x_{t-1},s_\mathbf{goal}), \\
\hat{s}_t &= \mathrm{CausalTrans}(a_{t-1}|x_{t-1},s_\mathbf{goal}).
\end{align}
Before being processed by the causal Transformer, states are encoded using relative position encoding following~\cite{sarkar2024gomaa}.

\noindent \textbf{Action Modeling Loss.}
We use the binary cross-entropy loss to guide the action prediction at each time step $t$ along the action-state sequence, as in previous work~\cite{sarkar2024gomaa}:
\begin{align}
\label{eq:action}
    \mathcal{L}_{\mathit{\text{Action}}} &= \sum_{t=0}^{N} -(y_t \log (P_t) + (1-y_t) \log (1-P_t))\nonumber,\\
    P_t &= \mathrm{Softmax}(\hat{a}_{t}),
\end{align}
where $N$ is the length of the random sequence. $P_t$ is the predicted probability of actions at $t$ when $x_{{t-1}}$ and $s_{\mathbf{goal}}$ are given as the input to the model, and $y_t = [y^{(1)}_t, y^{(2)}_t, \ldots, y^{(|\mathcal{A}|)}_t]$, indicates if an action from the $\mathcal{A}$ set is considered optimal at time step $t$. $y^{(j)}_t = 1$ if the $j$'th action leads agent closer to the goal, and $y^{(j)}_t = 0$ otherwise.

\noindent \textbf{State Modeling Loss.}
Without modifying the model architecture, we extend the causal Transformer from predicting only actions to jointly predicting both states and actions. We do so by adding a state modeling loss. Inspired by recent advances in world models~\cite{hafner2023mastering, bar2024navigation, zhou2024dino}, we propose to use the mean squared error between the predicted state representation and the ground truth state representation to guide the model learning:

\begin{equation}
\label{eq:state}
\mathcal{L}_{\mathit{\text{State}}} = \sum_{t=1}^{N-1} \left\|\hat{s}_t-s_t\right\|_2^2.
\end{equation}
Note that here the total sequence length is $N-1$ since we ignore the starting and ending state for prediction.

The final DM loss ($\mathcal{L}_{\mathit{\text{DM}}}$) is the combination of the action modeling loss ($\mathcal{L}_{\mathit{\text{Action}}}$) and the state modeling loss ($\mathcal{L}_{\mathit{\text{State}}}$):
\begin{equation}
\label{eq:dm}
\mathcal{L}_{\mathit{\text{DM}}}=\mathcal{L}_{\mathit{\text{Action}}} + \alpha \mathcal{L}_{\mathit{\text{State}}},
\end{equation}
where $\alpha$ is a scaling factor.

\subsection{Curiosity-Driven Exploration}
In this subsection, we first introduce the actor-critic pipeline for reinforcement learning, which serves as the foundation for our approach. Building on it, we then propose the curiosity-driven exploration guided by intrinsic rewards.

\noindent \textbf{Actor-Critic Reinforcement Learning Pipeline.}
During CE, the pre-trained causal Transformer is frozen. In this stage, we introduce an action prediction head as shown in Figure~\ref{fig:pipeline} (c), which includes an \textit{actor} $\pi_\theta\left(a|s_t\right)$ (policy network) and a \textit{critic} $v_\psi\left(s_{t}\right)$ (value function) optimized by Proximal Policy Optimization (PPO) following the actor-critic reinforcement learning pipeline~\cite{schulman2017proximal}. 
At each time step $t$, \textbf{(1)} the actor chooses an action $\hat{a}_t \sim \pi_\theta\left(a|s_t\right)$. \textbf{(2)} by executing $\hat{a}_t$, the state transits to $s_{t+1}$ and the reward $r^{CE}_{t}$ is obtained. \textbf{(3)} the value $v_\psi\left(s_{t}\right)$ denotes the expected total reward from state $s_t$ is calucated by the critic. \textbf{(4)} the actor and critic are updated under PPO. See \textit{supplementary materials} ($S2$) for details of the PPO agent. The agent operates on outputs of the causal Transformer and thus benefits from the action-state dynamics learned during DM.

\noindent \textbf{Curiosity-Driven Exploration.}
The most straightforward reward for goal-reaching RL problems is designed based on whether the goal is achieved or not. However, these rewards are usually very sparse, leading to insufficient feedback to learn efficiently and effectively~\cite{pathak2017curiosity}. Previous efforts proposed shaping the reward to be dense, and feasible to obtain during the process. In particular, existing methods in AGL used an extrinsic reward based on the distance between the current and the goal patch, to optimize the actor-critic action prediction head, which can be denoted as:

\begin{equation}
\resizebox{.9\hsize}{!}{$
    r^{\mathit{\text{ex}}}_{t} = \left\{
    \begin{array}
    {r@{\quad}l}
         \text{$1,$} & \text{if\>\>\>$|| p_{t+1} - p_{\mathbf{goal}} ||^2_2 < || p_{t} - p_{\mathbf{goal}} ||^2_2 $} \\
         \text{$-1,$}  & \text{if\>\>\>$|| p_{t+1} - p_{\mathbf{goal}} ||^2_2 > || p_{t} - p_{\mathbf{goal}} ||^2_2 \>\> \lor \>\> s_{t+1} \in \{x_{t}\}$},  \\
         \text{$2,$}  & \text{if\>\>\>$ s_{t+1} = s_{\mathbf{goal}}$}
    \end{array}
    \right.
    \label{eq:exreward}$}
\end{equation}
where $p_{t}$ denotes the position of patch $I_{t}$. The intuition behind the distance-based reward is \textit{learning from minimizing the distance}: The extrinsic reward encourages the agent when its actions reduce the distance to the goal, with the highest reward given upon successfully reaching the target location. Conversely, the agent will be penalized for moving away from the goal or revisiting previously explored states. However, when the goal location is \textit{unknown} (during inference), this reward becomes difficult to estimate and impacts the actor-critic network negatively.

To address this challenge, while encouraging exploration of the environment, we introduce intrinsic curiosity-driven rewards to AGL. Intrinsic rewards are designed following the intuition that an action is rewarded when it tends to explore an unexpected state. The degree of ``unexpectedness'' is measured based on the differences between state observation and prediction, the latter being built upon the state dynamics modeling in DM. Following the literature~\cite{pathak2017curiosity, ma2022elign}, we define the curiosity-driven intrinsic reward for AGL based on two different measurements:
\begin{itemize}
    \item \textbf{Mean squared error}. The reward is measured by the mean squared error between the predicted state representation ($\hat{s}_{t+1}$) and the observed state representation ($s_{t+1}$):
    \begin{equation}\label{eq:mse}
    r^{\mathit{\text{in}}}_{t}=\left\|\hat{s}_{t+1}-s_{t+1}\right\|_2^2.
\end{equation}
    \item \textbf{Cosine similarity}. The reward is measured by the cosine similarity between the two features.
    \begin{equation}
    r^{\mathit{\text{in}}}_{t}=-\mathrm{cos}(\hat{s}_{t+1},s_{t+1}).
\end{equation}
\end{itemize}

To combine the two sources of feedback (goal and the exploration), the final reward is a weighted sum of extrinsic and intrinsic rewards, weighted by a parameter $\beta$:
\begin{equation}
\label{eq:ce}
    r^{\mathit{\text{CE}}}_{t} = r^{\mathit{\text{ex}}}_{t} + \beta r^{\mathit{\text{in}}}_{t}.
\end{equation}
Note that before multiplying by $\beta$, we normalize intrinsic rewards to $\left[-1,1\right]$.

\subsection{Inference}
\label{secs:inference}
During inference, given a goal presented as one of the three modalities and a starting patch, GeoExplorer represents the goal and environment, and determines a trajectory of actions and states $x = \{s_{0}, \hat{a}_{0}, \ldots, s_{t}, \hat{a}_{t}, \ldots, s_{\mathit{M}} \}$ of length $\mathit{M}$ within a search budget $\mathcal{B}$ ($\mathit{M} \leq \mathcal{B}$) using the action prediction head. The actions are selected according to the argmax policy, \textit{i.e.}, the highest probability, for validation. For path visualization, we use stochastic policy, \textit{i.e.}, selecting action probabilistically, to show more possible paths.

\section{Data and Setup}
\label{sec:experiment}

\begin{figure}[t]
    \centering
    \includegraphics[width=0.95\linewidth]{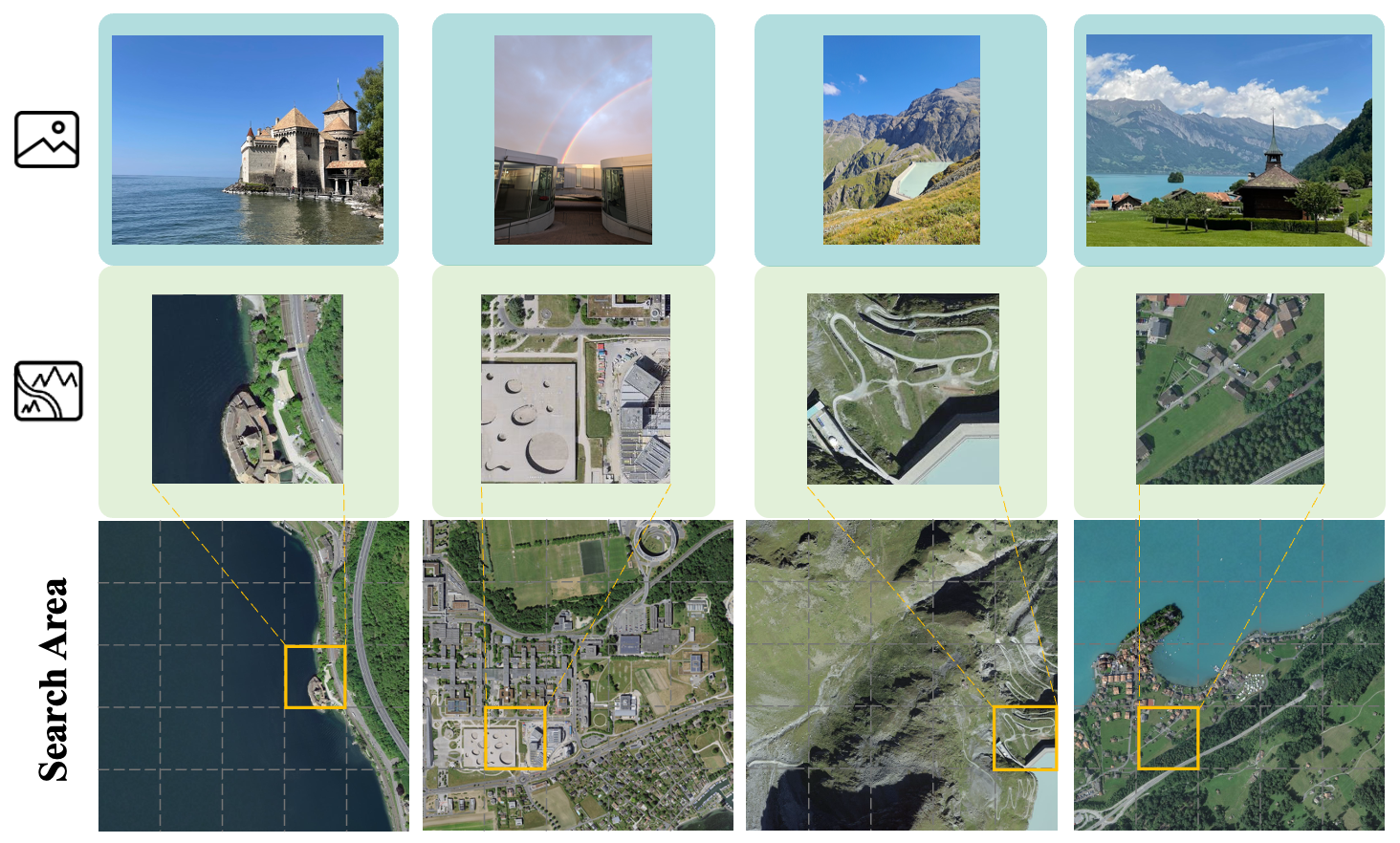}
    \caption{\textbf{Examples from the proposed SwissViewMonuments dataset with unseen targets}. From top to bottom: ground-level images, aerial goal patches, and search areas.} 
    \label{fig:dataset}
\end{figure}

\textbf{Datasets.}
We consider four AGL benchmarks:  
\textbf{The Masa dataset}~\cite{mnih2013machine} uses aerial views as goals. The dataset contains $1188$ images, with a split of $70\%$ for training and $15\%$ each for validation and test~\cite{pirinen2022aerial, sarkar2024gomaa}. Each image is regarded as a search area, and we randomly generate for each one 5 \{start, goal\} pairs to evaluate the model, resulting in $895$ configurations for testing. 
\textbf{The MM-GAG dataset}~\cite{sarkar2024gomaa} represents goals with different modalities such as text and ground-level images. It includes $73$ search areas. 
\textbf{The xBD dataset}~\cite{gupta2019creating} is a subset of the xBD dataset test set. Following previous work~\cite{sarkar2024gomaa}, $800$ images are kept for both before (xBD-pre) and after (xBD-disaster) disaster images. Note that to mimic real-world search-and-rescue missions, the goal is always represented with the pre-disaster image (xBD-pre), regardless of whether the xBD-pre or xDB-disaster is used as search areas.
Our proposed \textbf{SwissView dataset} (Figure~\ref{fig:dataset}) is constructed from Swisstopo's SWISSIMAGE $10$cm imagery, with two distinct components: SwissView100, which comprises $100$ images randomly selected from across the Swiss territory, thereby providing diverse natural and urban environment; and SwissViewMonuments, which includes $15$ images of atypical or distinctive scenes, such as landmarks and unseen landscapes, with corresponding ground level images. Since the targets in the latter are unique, it is used to assess models' generalization to unseen targets.
Detailed descriptions can be found in the \textit{supplementary materials} ($S1$). Following previous methods~\cite{pirinen2022aerial, sarkar2024gomaa}, and unless specified otherwise, GeoExplorer is only trained on the Masa dataset and evaluated on the others.

\noindent \textbf{Evaluation Metrics.} We use the \emph{success ratio} (SR) and \emph{step-to-the-goal} (SG) for evaluation. Within a search budget $\mathcal{B}$, if the agent reaches the goal, the attempt is considered successful. SR is defined as the ratio of successful localizations to the total number of attempts. SG is the Manhattan distance between the path-end and goal locations. The evaluation is carried out over different distances $\mathcal{C}$ from the start to the goal location. The results of SG are provided in the supplementary materials ($S4.7$).

\noindent \textbf{Evaluation Settings.} We evaluate GeoExplorer in four settings: 1) \textit{Validation}: we evaluate the model on the same dataset as it is trained. 2) \textit{Cross-domain transfer}: the model trained on the Masa dataset is evaluated on an unseen dataset using aerial view as goal modality. 3) \textit{Cross-modal generalization}: the goal is presented from different modalities. 4) \textit{Unseen target generalization}: to evaluate the models' adaptation to the unseen targets, we evaluate the model on our SwissView dataset, with aerial view and ground-level images as goals.

\noindent \textbf{Implementation Details.}
Following \cite{sarkar2024gomaa}, the search grid dimensions $X$ and $Y$ are set to $5$, leading to $25$ non-overlapping patches in a search area. The search budget $\mathcal{B}$ is set to $10$ and the start-to-goal distance is chosen in a set $\mathcal{C} \in \{ 4, 5, 6, 7, 8\}$. 
$\alpha$ (Eq.~\ref{eq:dm}) is set to 1 and $\beta$ (Eq.~\ref{eq:ce}) is set to $0.25$ according to parameter analysis (see in \textit{supplementary materials} $S4.1$). Eq.~\ref{eq:mse} is the default setting of intrinsic reward.
We use the pretrained CLIP vision (ViT-b-32) and text encoders~\cite{radford2018improving} as encoders for ground-level images and text descriptions. For aerial images, we use the Sat2Cap satellite encoder~\cite{dhakal2024sat2cap}, which is fine-tuned to align with the CLIP vision encoder using contrastive learning. The pre-trained Falcon-7B model~\cite{almazrouei2023falcon} is used as the causal Transformer in GeoExplorer, similarly to~\cite{sarkar2024gomaa}. The hyperparameter settings and model chosen follow the baselines to ensure a fair comparison. Detailed descriptions can be found in the \textit{supplementary materials} ($S2$).

\noindent \textbf{Baselines.} We compare GeoExplorer against:
\begin{itemize}
    \item \emph{Random policy}, which randomly selects an action at each time step; \emph{PPO policy~\cite{schulman2017proximal}} selects actions based on the current observation.
    \item \emph{Decision Transformer (DiT)}~\cite{chen2021decision}, an RL pipeline trained with offline optimal trajectories with randomly generated \{start, goal\} pairs via action modeling.
    \item \emph{AiRLoc}~\cite{pirinen2022aerial}, an RL-based model designed for unimodal AGL tasks. A goal-related sparse reward is used to guide model learning.
    \item \emph{GOMAA-Geo}~\cite{sarkar2024gomaa}, an RL-based method agnostic to multiple goal modalities for AGL tasks. It involves pretraining based on action modeling and an RL stage guided by distance-based extrinsic rewards.
\end{itemize}

\section{Results}
\subsection{Numerical Results in the Evaluation Settings}

\begin{table}[!t]
    \centering
    \small
    \caption{Numerical results  on the test set of the \textbf{Masa dataset}. The success ratio (SR) is reported with different start-to-goal distance $\mathcal{C}$. The results of the baselines are taken from~\cite{sarkar2024gomaa}.}
     \resizebox{0.96\columnwidth}{!}{
    \begin{tabular}{lccccc}
        \toprule
        Method & $\mathcal{C}=4$ & $\mathcal{C}=5$ & $\mathcal{C}=6$ & $\mathcal{C}=7$ & $\mathcal{C}=8$ \\
        \cmidrule(r){1-6} 
        Random policy & 0.1412 & 0.0584 & 0.0640 & 0.0247 & 0.0236 \\
        PPO policy~\cite{schulman2017proximal}& 0.1427 & 0.1775 & 0.1921 & 0.2269 & 0.2595 \\
        AiRLoc~\cite{pirinen2022aerial} & 0.1786 & 0.1561 & 0.2134 & 0.2415 & 0.2393 \\
        DiT~\cite{chen2021decision} & 0.2011 & 0.2956 & 0.3567 & 0.4216 & 0.4559  \\ 
        GOMAA-Geo~\cite{sarkar2024gomaa} & 0.4090 & 0.5056 & 0.7168 & 0.8034 & 0.7854 \\ 
        \textbf{GeoExplorer} & \textbf{0.4324} & \textbf{0.5318} & \textbf{0.8156} & \textbf{0.9229} & \textbf{0.9497}\\ 
        \bottomrule
    \end{tabular}
    }
    \label{tab: masa}
\end{table}

\begin{table}[t]
    \centering
    \small
    \caption{\textbf{Generalization across domain and goal modalities} on the \textbf{MM-GAG dataset}. The goal is presented as an aerial image (``I''), a ground-level image (``G''), or a text (``T''). Note that the models are only trained on the Masa dataset.}
    \resizebox{0.96\columnwidth}{!}{
    \begin{tabular}{clccccc}
        \toprule
        Goal & Method & $\mathcal{C}=4$ & $\mathcal{C}=5$ & $\mathcal{C}=6$ & $\mathcal{C}=7$ & $\mathcal{C}=8$  \\
        \cmidrule(r){1-7}
        \multirow{6}{*}{I} & Random policy & 0.1412 & 0.0584 & 0.0640 & 0.0247 & 0.0236\\
        & PPO policy~\cite{schulman2017proximal}&  0.1489 & 0.1854 & 0.1879 & 0.2176 & 0.2432\\
        & AiRLoc~\cite{pirinen2022aerial} & 0.1745 & 0.1689 & 0.2019 & 0.2156 & 0.2290\\
        & DiT~\cite{chen2021decision} & 0.2023 & 0.2856 & 0.3516 & 0.4190 & 0.4423\\ 
        & GOMAA-Geo~\cite{sarkar2024gomaa} & 0.4085 & 0.5064 & 0.6638 & 0.7362 & 0.7021\\ 
        & \textbf{GeoExplorer} & \textbf{0.4338} & \textbf{0.5415} & \textbf{0.7631} & \textbf{0.8369} & \textbf{0.8277}\\ 
        \multirow{2}{*}{G} & \cellcolor{mygrey}GOMAA-Geo~\cite{sarkar2024gomaa} & \cellcolor{mygrey}\textbf{0.4383} & \cellcolor{mygrey}\textbf{0.5150} & \cellcolor{mygrey}0.6808 & \cellcolor{mygrey}0.7489 &  \cellcolor{mygrey}0.6893\\ 
        & \cellcolor{mygrey}\textbf{GeoExplorer} & \cellcolor{mygrey}0.4308 & \cellcolor{mygrey}0.5138 & \cellcolor{mygrey}\textbf{0.7200} & \cellcolor{mygrey}\textbf{0.8246} & \cellcolor{mygrey}\textbf{0.7815}\\ 
        \multirow{2}{*}{T} & GOMAA-Geo~\cite{sarkar2024gomaa} &  0.4000 & \textbf{0.4978} & 0.6766 & 0.7702 & 0.6595\\
        & \textbf{GeoExplorer} & \textbf{0.4431} &  0.4892 & \textbf{0.7200} & \textbf{0.8062} & \textbf{0.7631}\\
        \bottomrule
    \end{tabular}}
    \label{tab: mmgag}
\end{table}

\noindent \textbf{Validation.} 
We first evaluate the models on the Masa dataset, \textit{i.e.}, the dataset used for training.
Table~\ref{tab: masa} demonstrates that GeoExplorer outperforms all other comparison methods on the Masa dataset, particularly for longer search paths. Specifically, it achieves a success ratio improvement of $0.0234$ when $\mathcal{C}=4$ and $0.1643$ when $\mathcal{C}=8$ compared to the baseline (GOMAA-Geo). Regarding reward design, approaches employing dense rewards, such as GOMAA-Geo or GeoExplorer demonstrate a significant advantage over the baseline utilizing sparse rewards (\textit{e.g.}, AiRLoc). This highlights the importance of dense and timely guidance in the AGL process. 
Interestingly, we observe higher performance for larger $\mathcal{C}$, even though we consider the task to be more challenging. This can be attributed to the limited variability in configurations in these scenarios, which makes it easier for the model to overfit to the training data distribution. (see \textit{supplementary materials} ($S5.1$) for details).

\noindent \textbf{Cross-Domain Transfer.} For an AGL agent, cross-domain transferability is crucial, as models are typically pre-trained on one dataset and subsequently deployed in an \textit{unseen environment}. Here, we provide the results of the models trained on the Masa dataset and then deployed on two other datasets, but again with aerial view as the goal modality. For  MM-GAG (Table~\ref{tab: mmgag}), when the goal is presented as an aerial view (the first block \emph{`Goal = I'}), GeoExplorer leads to the best overall performance, with a notable improvement of $0.0772$ on average. Similarly, the results for the xBD dataset (shown in Table~\ref{tab: xbd}) confirm the effectiveness of GeoExplorer in cross-domain transfer setting, with an average improvement of $0.0556$ on the xBD-disaster subsets. Note that this setting is extremely challenging since the goal is presented as pre-disaster images.

\begin{table}[!t]
    \centering
    \footnotesize
    \caption{\textbf{Cross-domain transfer} on the \textbf{xBD-disaster dataset}. Note that the models are only trained on the Masa dataset and the goal is always presented from the aerial view before the disaster.}
     \resizebox{0.96\columnwidth}{!}{
    \begin{tabular}{lccccc}
        \toprule
        Method & $\mathcal{C}=4$ & $\mathcal{C}=5$ & $\mathcal{C}=6$ & $\mathcal{C}=7$ & $\mathcal{C}=8$ \\
        \cmidrule(r){1-6} 
        Random policy & 0.1412 & 0.0584 & 0.0640 & 0.0247 & 0.0236\\
        PPO policy~\cite{schulman2017proximal}& 0.1132 & 0.1146 & 0.1292 & 0.1665 & 0.1953\\
        AiRLoc~\cite{pirinen2022aerial} & 0.1201 & 0.1298 & 0.1507 & 0.1631 & 0.1989\\
        DiT~\cite{chen2021decision} & 0.1012 & 0.2389 & 0.3067 & 0.3390 & 0.3543\\ 
        GOMAA-Geo~\cite{sarkar2024gomaa} & \textbf{0.4002} & 0.4632 & 0.6553 & 0.7391 & 0.6942\\
        \textbf{GeoExplorer} & 0.3975 & \textbf{0.5025} & \textbf{0.7185} & \textbf{0.8190} & \textbf{0.7923}\\
        \bottomrule
    \end{tabular}
    }
    \label{tab: xbd}
\end{table}

\noindent \textbf{Cross-Modal Generalization.} In real-world applications, the goal may be described using other modalities than aerial images.
Experimental results on the MM-GAG dataset presented in Table~\ref{tab: mmgag} suggest that with the same modality-specific encoders as used in GOMAA-Geo, GeoExplorer still shows better generalizability when the goal is described in ground-level images (second block  \emph{`Goal = G'}) and text (the third block  \emph{`Goal = T'}). With less direct information, GeoExplorer achieves performances comparable to the baseline on short paths, while significantly improving SR when the path is longer ($0.0690$ (ground-level images) and $0.0610$ (text) on average for $\mathcal{C}=\{6, 7, 8\}$).

\noindent \textbf{Unseen Target Generalization.} To assess the generalization ability of GeoExplorer and the baseline models, we design an \textit{unseen target} generalization experiment on the SwissViewMonuments dataset. Unlike previous datasets, where goal patches were randomly selected, this dataset has fixed goal locations. As suggested in Table~\ref{tab: unseen}, GeoExplorer exhibits an impressive generalization ability in localizing unseen targets, especially when the path is long (\textit{e.g.}, for $\mathcal{C}=6$, improving $0.1433$ when using aerial view and $0.1366$ when using ground-level images). 
More results from the SwissView dataset and evaluation using SG as metric are presented in \textit{supplementary materials} ($S4$).

\begin{table}[t]
    \centering
    \caption{\textbf{Unseen objects generalization ability} on the \textbf{SwissViewMonuments dataset}. $^{\star}$ corresponds to results obtained using the pretrained model~\cite{sarkar2024gomaa}.}
    \small
     \resizebox{0.95\columnwidth}{!}{
    \begin{tabular}{clccccc}
        \toprule
        & Method &  $\mathcal{C}=4$ & $\mathcal{C}=5$ & $\mathcal{C}=6$ & $\mathcal{C}=7$ & $\mathcal{C}=8$ \\
        \cmidrule(r){1-7} 
        \multirow{2}{*}{I} & GOMAA-Geo$^*$  & 0.4027 & 0.3833 & 0.6267 & 0.7278 & 0.7833 \\ 
        & \textbf{GeoExplorer} & \textbf{0.4133} & \textbf{0.5333} &  \textbf{0.7700} &  \textbf{0.8889} & \textbf{0.8833} \\
        \multirow{2}{*}{G} &  \cellcolor{mygrey}GOMAA-Geo$^*$  & \cellcolor{mygrey}0.4107 & \cellcolor{mygrey}0.3833 & \cellcolor{mygrey}0.6367 & \cellcolor{mygrey}0.7722 & \cellcolor{mygrey}0.7500 \\ 
        & \cellcolor{mygrey}\textbf{GeoExplorer} & \cellcolor{mygrey}\textbf{0.4160}  & \cellcolor{mygrey}\textbf{0.5167} &  \cellcolor{mygrey}\textbf{0.7733} & \cellcolor{mygrey}\textbf{0.8778} & \cellcolor{mygrey}\textbf{0.7833} \\
        \bottomrule
    \end{tabular}
    }
    \label{tab: unseen}
\end{table}

\begin{table}[t]
    \centering
    \small
    \caption{\textbf{Ablation study} on the Masa dataset on state modeling and reward design. ``MSE'' and ``COS'' denote intrinsic reward based on mean squared error or cosine similarity, respectively.}
    \resizebox{0.48\textwidth}{!}{
    \begin{tabular}{ccccp{0.90cm}p{0.90cm}p{0.90cm}p{0.90cm}p{0.90cm}}
        \toprule
        $\mathcal{L}_{\mathit{\text{Action}}}$ & $\mathcal{L}_{\mathit{\text{State}}}$ & $r^{\mathit{\text{ex}}}_t$ & $r^{\mathit{\text{in}}}_t$ & $\mathcal{C}=4$ & $\mathcal{C}=5$ & $\mathcal{C}=6$ & $\mathcal{C}=7$ & $\mathcal{C}=8$\\
        \cmidrule(r){1-9}
        \checkmark &  & \checkmark & & {0.4090} & {0.5056} & {0.7168} & {0.8034} & {0.7854}\\ 
        \checkmark & \checkmark & \checkmark &  & 0.3978 & 0.4939 & 0.7609 & 0.8413 & 0.8648\\
        \checkmark &  & \checkmark & MSE & 0.3888 & 0.4938 & 0.7408 & 0.8615 & 0.9140\\ 
        \checkmark & \checkmark & \checkmark & COS & 0.3955 & \textbf{0.5385} & 0.8134 & 0.9050 & 0.9352\\
         \checkmark & \checkmark & \checkmark & MSE & \textbf{0.4324} & 0.5318 & \textbf{0.8156} & \textbf{0.9229} & \textbf{0.9497}\\
        \bottomrule
    \end{tabular}}
    \label{tab: ablation}
\end{table}

\begin{figure*}[t]
    \centering
    \includegraphics[width=0.96\linewidth]{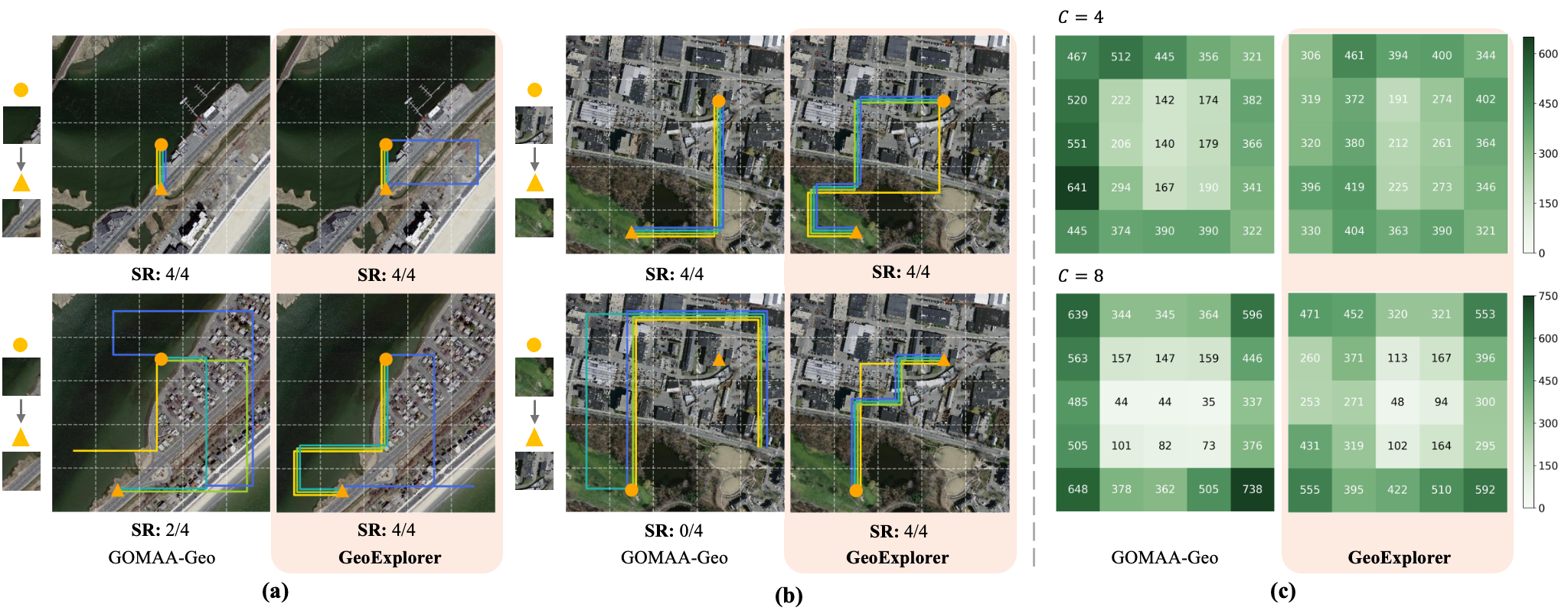}
    \caption{\textbf{GeoExplorer shows robust, diverse, and content-related exploration ability}. Given a pair of \{start ($\circ$), goal ($\triangle$)\} per search area, models generate four trials with stochastic policy (Sec.~\ref{secs:inference}), randomly shown in four different colors. (a) {Similar \{start, goal\} pairs with different distance}. Despite the difficulties of estimating distance, GeoExplorer's exploration ability maintains robust and diverse.
    (b) {Reversed \{start, goal\} pairs in the same search area}. GeoExplorer's exploration is adaptive and robust to goal content irrespective of the search direction.
    (c) Statistics of visited patches by all generated paths for both models when the distance $\mathcal{C}=4$ and $\mathcal{C}=8$ in Masa test set. Numbers denote the counts of visits per patch. GeoExplorer develops a more comprehensive exploration covering the search area.}
    \label{fig:path}
\end{figure*}

\subsection{Ablation Study}
We conduct ablation studies on the Masa dataset: first, we vary the training objectives in the DM stage to analyze the impact of the proposed Action-State Dynamics Modeling strategy (Eq.~\ref{eq:dm}). Comparing Row 2 with Row 1, by adding state modeling loss, SR improves $0.0277$ on average, which demonstrates the effectiveness of state transition modeling on action prediction, especially for longer distances. Adding intrinsic reward without state modeling also shows improvements ($0.0357$ on average, Row 3 and 1), which indicates that state prediction without explicit supervision, even though inaccurate, still provides useful information for exploration.
GeoExplorer with action-state dynamics modeling yields the best overall performance. With a learned state modeling, SR increases $0.0507$ in average (Row 5 and 3).
We also test different configurations of the reward design. Results show similar performance for the two reward designs, with an average SR difference of $0.0130$. We noticed a slight performance drop on short paths (Row 1 and 2/3). For shorter paths, the agent has limited steps to the goal, leaving less room for exploration, as it may lead to detours. Additionally, state modeling on short paths may be inaccurate due to limited history state observations.

\begin{figure}[t]
    \centering
    \includegraphics[width=0.97\linewidth]{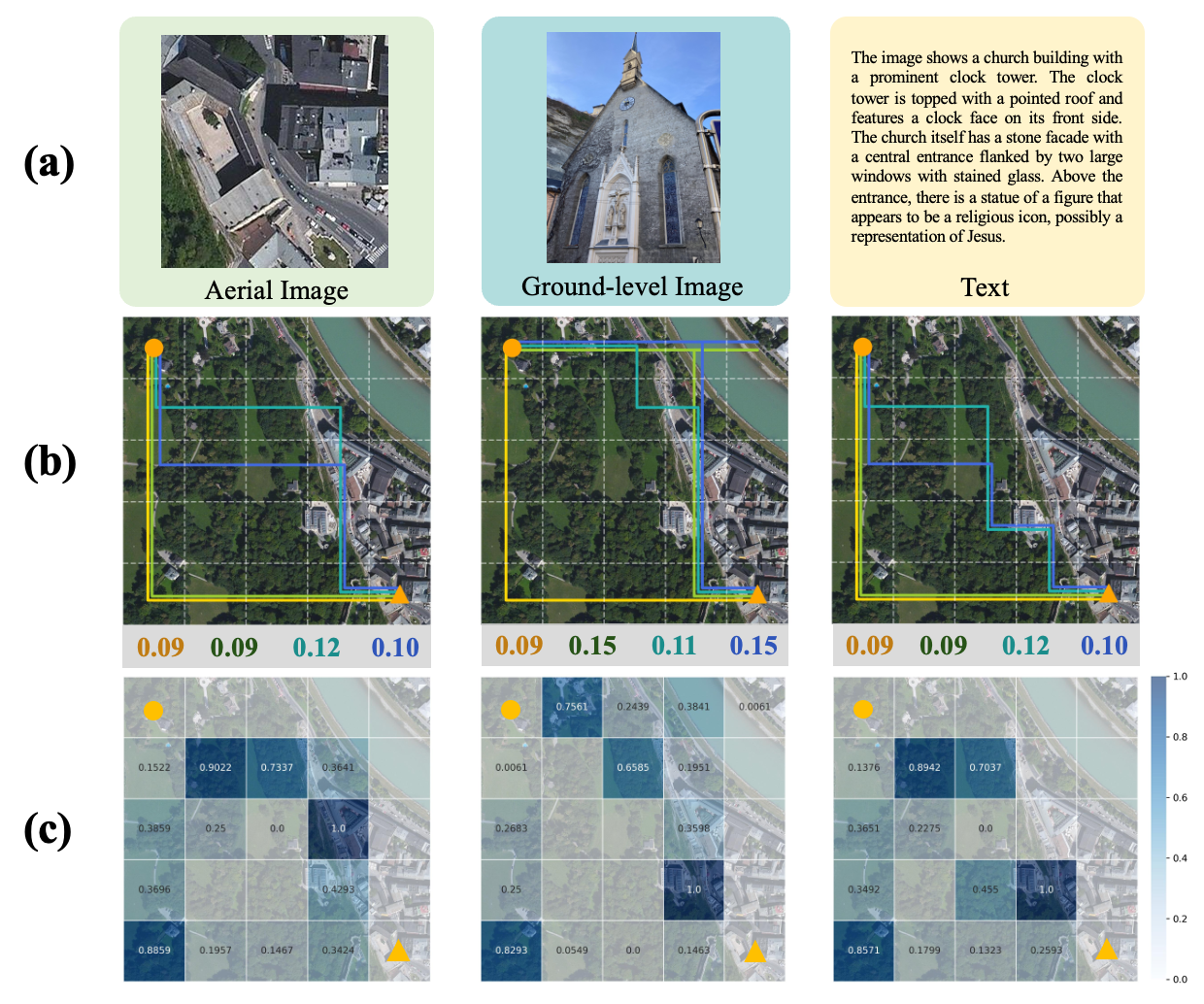}
    \caption{\textbf{Curiosity-driven intrinsic rewards provide dense, goal-agnostic, and content-related guidance to enhance the exploration ability of GeoExplorer.} Examples from the MM-GAG dataset targeting multiple goal modalities. (a) The goal of each example (column) is represented in different modalities. (b) Four trials, as well as average intrinsic rewards along the paths. (c) Average intrinsic reward from four trials at the patch level.}
    \label{fig:pattern}
\end{figure}

\subsection{Does Intrinsic Reward Improve Exploration?}
\noindent \textbf{Visualization of Exploration Ability.} To better understand the exploration capabilities of GeoExplorer and the baseline GOMAA-Geo, we visualize the generated paths in Figure~\ref{fig:path}. In panel (a), we compare \{start, goal\} pairs with similar semantic content from two different search areas. When the distance between the start and the goal is short and easy to estimate, GOMAA-Geo successfully reaches the goal with the shortest path (upper example). However, in the lower example, where visually similar start and goal patches are located in different parts of the image, GOMAA-Geo struggles to accurately localize the goal, as it finds it more challenging to estimate the potential distance. In contrast, GeoExplorer succeeds, leveraging its action-state modeling ability to differentiate between possible paths. In Figure~\ref{fig:path} (b), we reverse the \{start, goal\} patches to evaluate the model's robustness. GOMAA-Geo fails to maintain robustness, while GeoExplorer benefits from its exploration ability and successfully reaches the target irrespective of the search direction. The visited patch distribution shown in Figure~\ref{fig:path} (c) provides a more comprehensive observation of the exploration ability: when $\mathcal{C}=4$, only $20.08\%$ of the visited patches are located in the central part of the search aera for GOMAA-Geo, while GeoExplorer increases this ratio to $30.79\%$, indicating a more comprehensive exploration. Those findings further confirm that Geoexplorer shows robust, diverse, and content-related exploration ability.

\noindent \textbf{Analysis of Curiosity Patterns.} We provide an in-depth analysis of the curiosity-driven intrinsic reward. In Figure~\ref{fig:pattern} (b), we show four paths retrieved by our model and their corresponding average accumulated intrinsic reward (in the same colors as the path). As expected, we observe that trajectories visiting diverse image content (\textit{e.g.}, forest, city and river as in the green and blue trajectories) present higher average intrinsic rewards than those crossing forest patches only (\textit{e.g.}, the yellow trajectory). Figure~\ref{fig:pattern} (c) presents the curiosity scores per patch. Note that for visualization, we normalize the intrinsic rewards. Curiosity-reward is naturally dense and content-related. Patches with higher intrinsic reward turn out to be more ``interesting'', \textit{i.e.}, the semantic content of these patches can hardly be predicted from the surrounding patches. For instance, the patch with the highest score in the right column example depicts the transition from a forest to an urban environment. These results show that curiosity-driven rewards provide dense, goal-agnostic, and content-related guidance, enhancing the exploration ability of GeoExplorer.

\section{Conclusion}
\label{sec:conclusion}

Adaptive and active exploration is a step toward developing AGL in real-world search-and-recue operations.
In this paper, we propose GeoExplorer, an AGL agent that explores a search region with a combination of goal-oriented and \textit{curiosity-driven} guidance based on comprehensive \textit{environment modeling}. The curiosity-driven reward is goal-agnostic and provides dense and reliable feedback by leveraging the discrepancy between the state prediction and observation to encourage exploration. Extensive experiments and analyses on four AGL benchmarks demonstrate that GeoExplorer outperforms state-of-the-art methods and exhibits remarkable transferability to unseen targets and environments, highlighting its robust, generalizable, content-aware exploration ability driven by intrinsic curiosity.

\section*{Acknowledgement}
We thank the anonymous reviewers for their constructive and thoughtful comments. We thank Silin Gao, Valentin Gabeff, Chang Xu, Filip Dorm, Syrielle Montariol, and Sepideh Mamooler for providing helpful feedback on earlier versions of this work. We acknowledge the support from the CSC and EPFL Science Seed Fund, and the project “Transforming underwriting for commercial insurance with high-resolution remote sensing”, between EPFL and AXA Group Operations Switzerland AG. AB gratefully acknowledges the support of the Swiss National Science Foundation (No. 215390), Innosuisse (PFFS-21-29), the EPFL Center for Imaging, Sony Group Corporation, and a Meta LLM Evaluation Research Grant.

\clearpage
\clearpage
\newpage
\section*{Supplementary Materials}

\setcounter{section}{0}
\renewcommand{\thesection}{S\arabic{section}}
\setcounter{table}{0}
\renewcommand{\thetable}{S\arabic{table}}
\setcounter{figure}{0}
\renewcommand{\thefigure}{S\arabic{figure}}
\newcommand\blfootnote[1]{%
  \begingroup
  \renewcommand\thefootnote{}\footnote{#1}%
  \addtocounter{footnote}{-1}%
  \endgroup
}

The supplementary materials are organized as follows:
\begin{itemize}
    \item \textbf{Dataset details} (Section~\ref{ssec:dataset}).
    \item \textbf{Implementation details} (Section~\ref{ssec:implementation}).
    \item \textbf{Comparison of AGL and related tasks} (Section~\ref{ssec:task}).
    \item \textbf{Supplementary experiments}: parameter analysis, experiments on varying budget and larger grid size, evaluations using step-to-the-goal as metric, and supplementary results (Section~\ref{sec:suppexp}).
    \item \textbf{Supplementary analysis}: path analysis, additional visualization samples and failure case analysis (Section~\ref{sec:suppana}).
    \item \textbf{Discussions}: limitations and future work (Section~\ref{sec:disscusion}).
\end{itemize}

\section{Datasets Details}
\label{ssec:dataset}

\subsection{Massachusetts Buildings (Masa) Dataset}

\noindent \textbf{Data Collection.} The Massachusetts Buildings (Masa) dataset~\cite{mnih2013machine} consists of 1188 high resolution images of the Boston area. Building footprint annotations were obtained by rasterizing data from the OpenStreetMap project.
\noindent \textbf{Dataset Composition.} The dataset is split in $70\%$ for training ($832$ images), $15\%$ for testing and evaluation ($178$ for validation and $179$ for testing)~\cite{pirinen2022aerial, sarkar2024gomaa}. Each image, or search area, is structured as a $5$ $\times$ $5$ grid of search cells, with $300$ $\times$ $300$ pixels per grid cell. During training, data augmentation is applied through top-right and left-right flipping, and each search area allows for $25$ start positions with $24$ possible goal locations, leading to approximately 2 million unique training trajectories. For testing and validation, only a fixed configuration is randomly selected per start-to-goal distance and per search area, ensuring $895$ fixed test trajectories.
\subsection{MM-GAG Dataset}

The MM-GAG dataset~\cite{sarkar2024gomaa} was constructed to address the limitations of existing datasets for Active Geo-localization (AGL), which often lack precise coordinate annotations and meaningful goal representations across diverse modalities. 

\noindent \textbf{Multi-Modal Goal Representations.} Unlike many existing datasets that focus solely on aerial-to-aerial or aerial-to-ground localization, MM-GAG introduces multi-modal goal representations: \begin{itemize}
    \item Aerial Imagery,
    \item Ground-Level Imagery, 
    \item Natural Language Descriptions.
\end{itemize}

\noindent \textbf{Data Collection.} The MM-GAG dataset was built by collecting high-quality geo-tagged images from smartphone devices across diverse locations. Images have been filtered, resulting in $73$ distinct search areas. Note that through the link provided in the original paper\footnote{\hyperlink{MM-GAG}{https://huggingface.co/datasets/MVRL/MM-GAG/tree/main}}, we only find 65 search areas. To ensure fair comparison, we evaluate the proposed method and the pretrained baseline model provided in the original paper in Section~\ref{sec:suppexp}. For each of the $73$ ground-level images, high-resolution satellite image patches were retrieved at $0.6$m per pixel resolution. From these patches, $5$ $\times$ $5$ search grids with $256$ $\times$ $256$ pixels per grid cell were constructed. To generate textual goal descriptions, each ground-level image was automatically captioned using LLaVA-7B ~\cite{liu2023visual}. The captioning prompt was carefully designed to ensure concise and relevant descriptions.

\noindent \textbf{Dataset Composition.} Trajectories are selected by randomly sampling start and goal locations within each of the 73 search areas, ensuring a diverse range of search scenarios. For each area, five \{start, goal\} pairs are chosen for every predefined distance category, resulting in a total of 365 evaluation trajectories per start-to-goal distance.

\subsection{xBD Dataset}
The xBD dataset~\cite{gupta2019creating} is a large-scale aerial imagery dataset designed for disaster analysis. It contains images captured before (xBD-pre) and after (xBD-disaster) various natural disasters such as wildfires, floods, and earthquakes.

\noindent \textbf{Data Collection.} Imagery for the original xBD dataset was sourced from the Maxar/DigitalGlobe Open Data Program\footnote{\hyperlink{DigitalGlobe}{https://www.digitalglobe.com/}}, which provides high-resolution satellite images for major crisis events. The dataset includes imagery from 19 natural disasters across $45,361.79$ km² of affected areas. A total of $22,068$ images were collected, covering $850,736$ human-annotated building polygons. 

\begin{figure*}[t]
    \centering
    \includegraphics[width=0.95\linewidth]{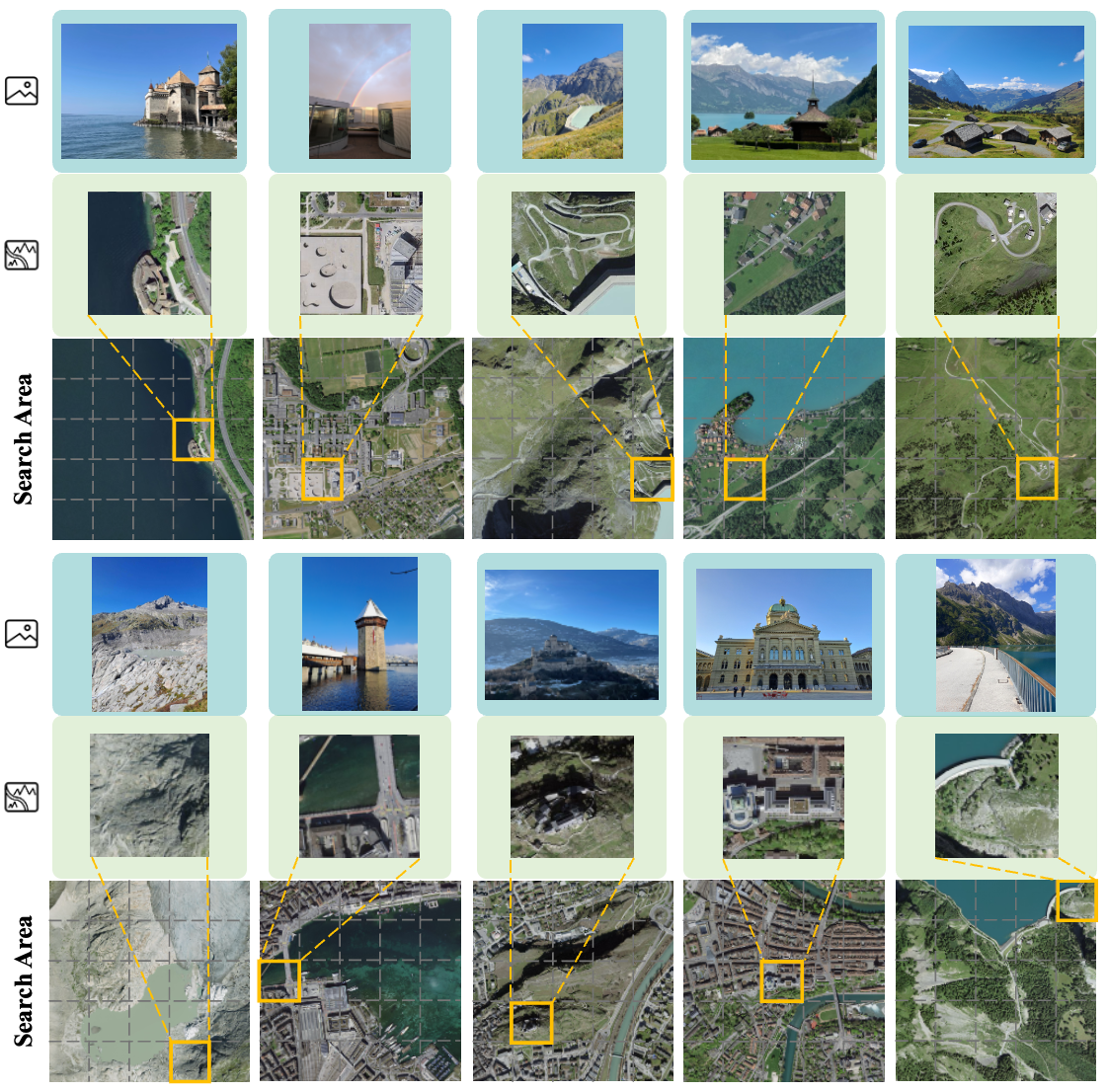}
    \caption{\textbf{Additional examples from the proposed SwissViewMonuments dataset} with unseen targets. From up to down: goal presented as ground-level images, goal presented in aerial view, and the search areas.} 
    \label{fig:swissview}
\end{figure*}

\noindent \textbf{Dataset Composition.} For the specific task of AGL, 800 bitemporal search areas (one image before and one image after the disaster) have been selected from the original dataset. Each search area corresponds to a $5$ $\times$ $5$ search grid with $300$ $\times$ $300$ pixels per grid cell. Trajectories are obtained by randomly sampling $5$ pairs of start and goal location per start-to-goal distance, resulting in $4000$ evaluation trajectories per start-to-goal distance.

\subsection{SwissView Dataset} 

\begin{figure}[t]
    \centering
    \includegraphics[width=0.95\linewidth]{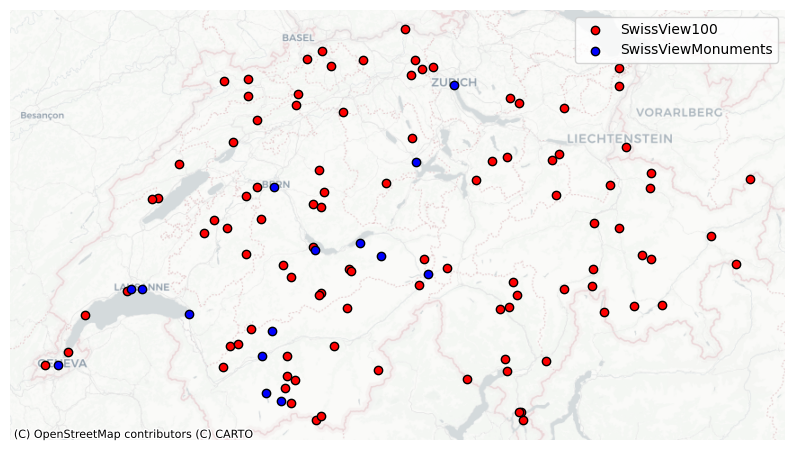}
    \caption{Geographic distribution of the images from the SwissView dataset. Red points indicate the locations of the 100 randomly sampled images from the SwissView100 subset, while blue points represent the images from the SwissViewMonuments subset.} 
    \label{fig:swissview_distrib}
\end{figure}

\paragraph{Data Collection.}  The SwissView dataset consists in two complementary components: SwissView100 and SwissViewMonuments. For SwissView100, a total of 100 images were randomly sampled across the entire territory of Switzerland, sourced from Swisstopo's SWISSIMAGE 10 cm database\footnote{\hyperlink{SwissTopo}{https://www.swisstopo.admin.ch/fr/orthophotos-swissimage-10-cm}}. The spatial distribution of the images is provided in Figure~\ref{fig:swissview_distrib}. The original images, with a spatial resolution of $0.1$ meters per pixel and dimensions of $10,000$ $\times$ $10,000$ pixels, were downsampled to a resolution of $0.6$ meters per pixel, resulting in $1500$ $\times$ $1500$ pixels images. These downsampled images were subsequently partitioned into $5$ $\times$ $5$ patches, each measuring $300$ $\times$ $300$ pixels. 
For SwissViewMonuments, the procedure is identical, if only for the choice of images. For this part of the dataset, 15 specific areas of Switzerland have been carefully selected for their atypical constructions or landscapes. We consider targets from two categories: 1) \textit{object uniqueness}: landmarks or localizable architectures; 2) \textit{location and scene uniqueness}: unseen scene classes. The resulting dataset, for example, includes images from remarkable buildings such as cathedrals and castles, and rare landscapes like glaciers.
Besides the aerial view of the search area, we also provide the corresponding ground-level images and its location associated to the aerial view. A few examples from the SwissViewMonuments dataset are shown in Figure~\ref{fig:swissview}. 

\paragraph{Dataset Statistics.} The location of the selected areas is also given in Figure~\ref{fig:swissview_distrib}. Samples from the SwissView100 dataset (in red) are distributed among the region and samples from the SwissViewMonuments (in blue) are chosen from cities, tourist attractions and nature reserves.

\paragraph{Dataset Composition.} To generate trajectories for the SwissView100 subset, we followed a similar approach to other datasets by randomly generating $5$ \{start, goal\} pairs for each trajectory and each start-to-goal distance, resulting in $500$ trajectories per distance considered. In contrast, for the SwissViewMonuments subset, we provide $25$ shifted aerial views for each aerial-ground pair ($15 \times 25$ samples in total), fixing the goal at all positions in a $5\times5$ grid. For each specified start-to-goal distance $\mathcal{C} \in \{4, 5, 6, 7, 8\}$, we randomly select one starting point per sample that satisfies the given distance relative to the fixed goal. Samples with goal positions that do not permit any valid starting point at a given distance are excluded from the evaluation for that distance. This process results in $\{375, 360, 300, 180, 60\}$ configurations for distances $\mathcal{C}$ in $\{4, 5, 6, 7, 8\}$, respectively.

\section{Implementation Details}
\label{ssec:implementation}
This section provides an overview of the implementation details, including the pretrained models used for text, ground-level, and aerial image encoding, as well as the causal transformer used for action-state modeling. Note that apart from the action-state dynamics modeling and curiosity-driven component of our model, the implementation and training parameters remain consistent with those outlined in the work of Sarkar \textit{et al.}~\cite{sarkar2024gomaa}, which serves as the baseline for our study.

\paragraph{Text and Ground-level Image Encoders.} To encode text descriptions and ground-level images of the goal, we use the pretrained encoders from the CLIP model~\cite{radford2018improving}, with the same pretrained weights used in~\cite{sarkar2024gomaa}, which are available on Hugging Face\footnote{\hyperlink{Clippretrain}{https://huggingface.co/openai/clip-vit-base-patch32}}. Specifically, the vision encoder is a Vision Transformer (ViT-b-32), and the text encoder is based on the BERT architecture, both of which are aligned in a shared multimodal embedding space through contrastive learning. These encoders remain frozen during training of the GeoExplorer model.

\paragraph{Aerial Image Encoder.} The aerial images are processed with the Sat2Cap satellite encoder~\cite{dhakal2024sat2cap}, which is fine-tuned to align its feature representations with the CLIP embedding space. The alignment is performed using contrastive learning with the InfoNCE loss~\cite{oord2018representation}, leveraging a large-scale dataset of paired aerial and ground-level images. Note that the CLIP image encoder remains frozen during this finetuning of the aerial image encoder. This alignment ensures that the features extracted from the aerial images share the same representation space as the features from the text descriptions and ground-level images. We use the same pretrained weights for Sat2Cap as the reference work~\cite{sarkar2024gomaa}, which can be found on Hugging Face\footnote{\hyperlink{Sat2Cap}{https://huggingface.co/MVRL/Sat2Cap}}.

\paragraph{Causal Transformer.}
For the Causal transformer used for sequential action and state prediction, we employ a pre-trained Falcon-7B model~\cite{almazrouei2023falcon}. The pretrained weights can be found on Hugging Face\footnote{\hyperlink{Falcon7B}{https://huggingface.co/openai/clip-vit-base-patch16}}. We follow the multi-modal projection layer introduced in GOMAA-Geo~\cite{sarkar2024gomaa}, to align the visual and language modalities into the latent space of the Falcon-7B model. Additionally, relative position encodings, measured with respect to the top-left position of the image, are incorporated into each state representation, allowing the model to encode spatial relationships between observed aerial images. Note that the primary distinction from the work of Sarkar et al.~\cite{sarkar2024gomaa} is that, in our approach, both states and actions are predicted sequentially, rather than solely predicting actions. However, the overall structure remains unchanged. The Causal transformer is trained using a learning rate of $1e-4$, a batch size of $1$, and the Adam optimizer over $300$ epochs. We followed the settings in GOMAA-Geo~\cite{sarkar2024gomaa} to ensure a fair comparison.

\begin{figure}[t]
    \centering
    \includegraphics[width=0.55\linewidth]{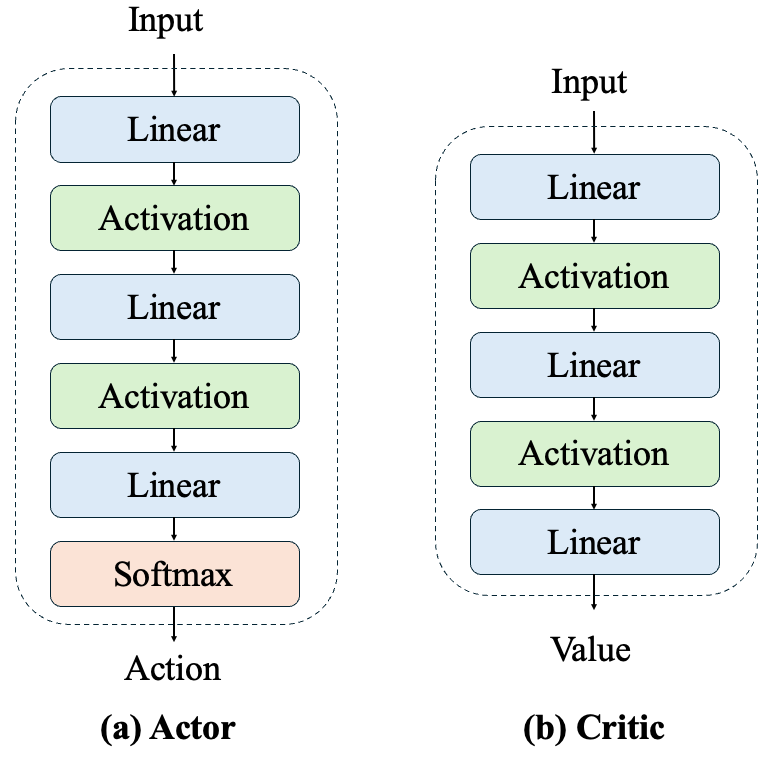}
    \caption{Model architecture of the actor-critic network (action prediction head).} 
    \label{fig:actorcritic}
\end{figure}

\paragraph{PPO.}

During the Curiosity-Driven Exploration (CE) phase, we introduce an action prediction head on top of the frozen pretrained Causal Transformer. Since the Causal Transformer alone does not inherently model decision policies, the addition of the action prediction head is crucial, as it allows the system to explicitly learn a mapping from the learned state representations to the concrete actions to take. The action prediction head is implemented using an actor-critic framework and optimized using the Proximal Policy Optimization (PPO)~\cite{schulman2017proximal}. This framework consists of an actor network responsible for policy learning, $\pi_\theta(a|s_t)$ and a critic network $v_\psi(s_t)$ that evaluates the expected total reward from state $s_t$. As shown in Figure~\ref{fig:actorcritic}, the actor and critic networks are implemented as Multi-Layer Perceptrons (MLPs) with three hidden layers. Each hidden layer is followed by a \texttt{tanh} activation function. The final layer of the actor network includes a \texttt{softmax} activation to output a probability distribution over actions, ensuring valid action selection. The critic network outputs a single scalar value representing the estimated value function.

At each time step $t$ (which specifies the time index in $\left[0, T\right]$), with the state representation $s_t$, the learning process of PPO can be described as:
\begin{enumerate}[label=(\textbf{\arabic*})]
    \item The actor chooses an action $\hat{a}_t \sim \pi_\theta\left(a|s_t\right)$, where $a \in \mathcal{A}$ is an action from avaliable action set.
    \item The agent executes the action in the environment and obtains the new state $s_{t+1}$ and reward $r^{CE}_{t}$.
    \item The \textit{value}, i.e. the total reward from state $s_t$ is calculated by the critic $v_\psi(s_t)$.
    \item The \textit{return}, i.e. the total reward from taking action $\hat{a}_t$ from state $s_t$ is calculated: \begin{align}
        q(s_t,\hat{a}_t) &= \sum_{t^{\prime}=t}^T E_{\pi_\theta}\left[r^{CE}_{t^{\prime}}\mid s_t, \hat{a}_t\right] \\ \notag
        &=r^{\mathit{\text{CE}}}_t + \gamma r^{\mathit{\text{CE}}}_{t+1} + ... \\
        & + \gamma^{T-t+1} r^{\mathit{\text{CE}}}_{T-1} + \gamma^{T-t}v_\psi(s_{T}). \notag
    \end{align}
    \item The advantage function $A_t$ is calculated to estimate how much better (or worse) an action $\hat{a}_t$ is compared to the average performance expectation at state $s_t$: \begin{equation}
        A_t = q(s_t,\hat{a}_t) - v_\psi(s_t).
    \end{equation}
    \item The actor and critic are updated.
\end{enumerate}

We update both the actor and critic networks using the PPO loss function, which consists of three main components: \begin{itemize}
    \item Actor loss (clipped surrogate objective), which compares the probabilities from the old and the updated policies and will constrain the policy change in a small range: \begin{equation}
        \mathcal{L}_{\text{Actor}}= \min \left\{\frac{\pi_\theta}{\pi_{\theta,\text{old}}}A_t, \: \text{clip}\left(1-\epsilon, 1+\epsilon, \frac{\pi_\theta}{\pi_{\theta,\text{old}}}\right)A_t\right\}. 
    \end{equation}
    \item Critic loss, which measures how well the model predicts the expected reward:\begin{equation}
        \mathcal{L}_{\text{Critic}}=(v_\psi(s_t)-q(s_t,\hat{a}_t))^2.
    \end{equation}
    \item Entropy regularization, which encourages policy exploration by preventing premature convergence to suboptimal policies:\begin{equation}
        \mathcal{H}\left[\pi_\theta\right](s_t)=-\sum_a \pi_\theta(a|{s_t}) \log \pi_\theta(a|{s_t}).
    \end{equation} 
\end{itemize}
The final loss function used for optimization is: \begin{equation}
    \mathcal{L}_\text{PPO}=\mathbb{E}\left[-\mathcal{L}_{\text{Actor}}+ \omega \mathcal{L}_{\text{Critic}} + \rho \mathcal{H}\right],
\end{equation}
where $\omega$ and $\rho$ are hyperparameters controlling the balance between policy learning, value estimation, and exploration.

We choose the hyperparameters to be consistent with the baseline model \cite{sarkar2024gomaa}. The learning rate is set to $1e-4$ and the batch size is $1$. The model is trained for $300$ epochs using the Adam optimizer. The values for the hyperparameters $\alpha$ and $\beta$ are set to $0.5$ and $0.01$, respectively. The clipping ratio $\epsilon$ is chosen to be $0.2$ and the discount factor $\gamma$ is set to $0.99$ for all experiments. As in \cite{sarkar2024gomaa}, we copy the parameters of $\pi_\theta$ onto $\pi_{\theta, \text{old}}$ every 4 epochs of policy training. For CE stage, patches are resized to $224\times224$ and normalized.

\section{Comparison of AGL and Related Tasks}
\label{ssec:task}
The following section provides a comparison of four related tasks: Active Geo-Localization, Visual Geo-Localization, Cross-View Geo-Localization, and Visual Navigation. It highlights their characteristics and differences, as well as associated challenges.

\paragraph{Active Geo-Localization.} Active Geo-Localization aims at locating a target by exploring an environment using a sequence of visual aerial observations~\cite{pirinen2022aerial}. This task is especially important in applications such as search-and-rescue~\cite{sarkar2024gomaa, pirinen2022aerial}, where efficient exploration is crucial for success. Unlike Visual Geo-Localization~\cite{zamir2016introduction}, where localization relies solely on unique and static observations, AGL involves movement of an agent to refine position estimates and ultimately reach the goal. Generally, the goal can be specified with different modalities, such as images or text descriptions~\cite{sarkar2024gomaa,pirinen2022aerial}. Reinforcement learning is often used to define the agent’s exploration strategy, guiding it towards the predefined target.

\paragraph{Visual Geo-Localization.} The task of Visual Geo-Localization~\cite{zamir2016introduction} is linked to the task of Active Geo-Localization in the sense that both aim at determining a location based on a given image. However, while Active Geo-Localization uses an agent to explore the environment to refine its position and reach the goal, visual geo-localization depends on single inputs, such as images or video frames, \textit{without the need for an agent to move}~\cite{vivanco2023geoclip, astruc2024openstreetview}. The input image’s location is estimated by comparing the observed images to an existing database of geotagged images, often leveraging image-retrieval techniques~\cite{berton2022rethinking, berton2022deep}. Visual geo-localization can operate in various settings, from small-scale areas like specific streets~\cite{astruc2024openstreetview} to large urban environments~\cite{berton2022rethinking}, depending on the breadth of the dataset used. Common applications include mobile device localization~\cite{chen2011mobile} or autonomous vehicles using street-view data~\cite{doan2019scalable}. 

\paragraph{Cross-View Geo-Localization.} The task of cross-view geo-localization aims at localizing a ground-level image by retrieving its corresponding geo-tagged aerial view~\cite{workman2015wide, workman2015wide, zhu2021vigor, zhang2024geodtr+}. Especially, fine-grained cross-view geo-localization~\cite{shi2023boosting, xia2025fg, shore2025peng, fervers2023uncertainty, wang2023fine} requires estimating the 3 Degrees of Freedom (DoF) pose of a query ground-level image on an aerial image, which is similar to the AGL setting. Although related, cross-view geo-localization requires \textit{full access} to the search area to perform matching, while AGL \textit{only provides the agent with partial visibility to the search area from the outset} and performs observation only along the exploration trajectory.

\paragraph{Visual Navigation.} Visual navigation~\cite{bar2024navigation, khanna2024goat, shah2023vint} is similar to Visual Geo-Localization as both tasks involve an agent exploring its environment to reach a predefined goal. However, unlike Visual Geo-Localization, which functions in aerial, \textit{i.e.}, bird-eye-view environments, visual navigation typically operates in a ground-level environment. Despite the similarities, Active Geo-Localization presents its unique challenges compared to Visual Navigation. One key difference is that, in Active Geo-Localization, \textit{the goal may not be visible to the agent in advance or even presented in different modalities from the agent observation, which introduces a level of uncertainty and complexity}. Moreover, the environment may change abruptly between two actions, as the agent can quickly transition from one type of terrain to another (such as moving from an urban region to a wooded area) due to the larger spatial scope of observations at each step. In contrast, the environment in which the agent operates is more localized in visual navigation, which allows for more accurate location estimation and easier navigation.

\paragraph{Vision-Language Navigation.} 
Vision-language navigation (VLN)~\cite{wang2024towards, gu2022vision}, especially Aerial VLN~\cite{liu2023aerialvln, fan2022aerial}, is linked to AGL as both tasks provide multimodal guidance to the agent to reach a goal in an environment. Unlike AGL, VLN performs the navigation in a continuous space in terms of both observation and action, which poses an additional challenge. However, AGL also has its unique challenges compared to VLN. Since VLN assumes that the instructor knows the goal's location, the natural language instructions are detailed throughout the navigation and typically correspond to the agent's actions (e.g.,``turn right''). In AGL, the only guidance provided is the goal information, making the setting more challenging due to the sparse reward and accumulated errors in a goal-reaching reinforcement learning context. Moreover, the mainstream methods of VLN are based on sequence modeling and prediction, which differs from the RL pipeline in AGL.

\section{Supplementary Experiments}
\label{sec:suppexp}
\subsection{Parameter Analysis}
\paragraph{The impact of loss weight $\alpha$.} We use loss weight $\alpha$ to balance the contribution of action modeling loss and state modeling loss. To evaluate the effectiveness of different ablations, we randomly generate the action-state trajectories with the optimal actions for each time step on the test set of the Masa dataset, following the steps described in Section 3.3 of the main paper. Then, we evaluate the models to predict optimal actions at each step of the trajectory. We use random seeds to ensure the same trajectories are tested for all the methods in a test, and we evaluate the models on 5 different tests. The action prediction accuracy for each test as well as the average prediction among 5 tests are reported in Table~\ref{tab:alpha_weight}. Among most of the tests except test 1, adding state modeling loss leads to a better action prediction performance, which confirms the fact that state and action transitions are inherently interconnected and dynamically influencing each other. When $\alpha=1$, the model achieves best overall performance with an improvement of $0.0271$ over the baseline with only the action modeling loss ($\alpha=0$).

\paragraph{The impact of reward weight $\beta$.} We also control the impact of intrinsic reward on the final reward by using reward weights $\beta$. Results in Table~\ref{tab:beta_weight} suggest a good balance should be achieved between the goal-oriented extrinsic reward, which directs the agent to the goal and the curiosity-driven intrinsic reward, which encourages the agent to explore the environment. The empirical results show that when $\beta=0.25$, the agent balances the guidance from extrinsic goal and intrinsic curiosity best.

\begin{table}[t]
    \centering
    \normalsize
    \caption{\textbf{Parameter analysis of loss weight $\alpha$} on the test set of the Masa dataset. The action prediction accuracy is reported for the DM stage. $\alpha=0$ denotes the baseline with action modeling loss only.}
    \resizebox{0.45\textwidth}{!}{
    \begin{tabular}{ccccccc}
        \toprule
        $\alpha$ & Test 1 & Test 2 & Test 3 & Test 4 & Test 5 & Average\\
        \cmidrule(r){1-7}
        \rowcolor{mygrey} 0 & \textbf{0.6056} & 0.1883 & 0.0838 & 0.6953 & 0.1429 & 0.3432 \\
        0.5 & 0.5162 & 0.1708 & \textbf{0.1133} & 0.7034 & 0.1429 & 0.3293\\
        1 & 0.5777 & \textbf{0.2602} & 0.0407 & \textbf{0.8299} & 0.1429 & \textbf{0.3703}\\
        2 & 0.5687 & 0.2179 & 0.0064 & 0.8055 & \textbf{0.1524} & 0.3502\\
        \bottomrule
    \end{tabular}}
    \label{tab:alpha_weight}
\end{table}

\begin{table}[t]
    \centering
    \small
    \caption{\textbf{Parameter analysis of reward weight $\beta$} on the test set of the Masa dataset. $\beta=0$ denotes the baseline with extrinsic reward only.}
    \resizebox{0.4\textwidth}{!}{
    \begin{tabular}{cccccc}
        \toprule
        $\beta$ & $\mathcal{C}=4$ & $\mathcal{C}=5$ & $\mathcal{C}=6$ & $\mathcal{C}=7$ & $\mathcal{C}=8$  \\
        \cmidrule(r){1-6}
        \rowcolor{mygrey} 0 &  0.3978 & 0.4939 & 0.7609 & 0.8413 & 0.8648\\
        0.25 & \textbf{0.4324} & \textbf{0.5318} & \textbf{0.8156} & \textbf{0.9229} & 0.9497 \\ 
        0.5 & 0.3597 & 0.4849 & 0.7687 & 0.9073 & \textbf{0.9587} \\
        1 & 0.3988 & 0.4983 & 0.7542 & 0.9028 & 0.9352 \\ 
        \bottomrule
    \end{tabular}}
    \label{tab:beta_weight}
\end{table}

\begin{table*}
    \centering
    \fontsize{6pt}{7pt}\selectfont
    \caption{\textbf{Cross-domain transfer} on the \textbf{xBD-pre and xBD-disaster datasets}. Note that the models are only trained on the Masa dataset and the goal is always presented from the aerial view before the disaster for both datasets.}
    \resizebox{0.99\textwidth}{!}{
    \begin{tabular}{lcccccccccc}
        \toprule
        & \multicolumn{5}{c}{Evaluation using xBD-pre Dataset} & \multicolumn{5}{c}{$\>\>\>\>\>\>$Evaluation using xBD-disaster Dataset} \\
        \cmidrule(r){1-11}
        Method & $\mathcal{C}=4$ & $\mathcal{C}=5$ & $\mathcal{C}=6$ & $\mathcal{C}=7$ & $\mathcal{C}=8$ & $\mathcal{C}=4$ & $\mathcal{C}=5$ & $\mathcal{C}=6$ & $\mathcal{C}=7$ & $\mathcal{C}=8$\\
        \cmidrule(r){1-6} \cmidrule(l){7-11}
        Random policy & 0.1412 & 0.0584 & 0.0640 & 0.0247 & 0.0236 & 0.1412 & 0.0584 & 0.0640 & 0.0247 & 0.0236\\
        PPO policy~\cite{schulman2017proximal}& 0.1237 & 0.1262 & 0.1425 & 0.1737 & 0.2075 & 0.1132 & 0.1146 & 0.1292 & 0.1665 & 0.1953\\ 
        AiRLoc~\cite{pirinen2022aerial} & 0.1191 & 0.1254 & 0.1436 & 0.1676 & 0.2021 & 0.1201 & 0.1298 & 0.1507 & 0.1631 & 0.1989\\
        DiT~\cite{chen2021decision} & 0.1132 & 0.2341 & 0.3198 & 0.3664 & 0.3772 & 0.1012 & 0.2389 & 0.3067 & 0.3390 & 0.3543\\ 
        GOMAA-Geo~\cite{sarkar2024gomaa} & 0.3825 & 0.4737 & 0.6808 & 0.7489 & 0.7125 & \textbf{0.4002} & 0.4632 & 0.6553 & 0.7391 & 0.6942\\
        \textbf{GeoExplorer} & \textbf{0.3973} & \textbf{0.4990} & \textbf{0.7328} & \textbf{0.8390} & \textbf{0.8363} & 0.3975 & \textbf{0.5025} & \textbf{0.7185} & \textbf{0.8190} & \textbf{0.7923}\\
        \bottomrule
    \end{tabular}}
    \label{tabs: xbd}
\end{table*}

\subsection{Results on the xBD dataset}
We present the supplementary results on the xBD dataset in Table~\ref{tabs: xbd}. The results show a small performance gap of GeoExplorer between the two subsets ($0.0149$ on average), indicating the generalization ability of the model. 

\subsection{Results on the SwissView100 dataset}
The proposed SwissView dataset has two subsets: SwissViewMonuments for unseen target generalization evaluation and SwissView100 for cross-domain transfer evaluation. We present the results from the former setting in the main paper and the latter one in this section. As stated before, the test setting for SwissView100 dataset is the same as the cross-domain transfer setting on the MM-GAG Aerial and xBD dataset: the model is only trained on the Masa dataset and the goal is presented in aerial view. As shown in Table~\ref{tab:swissview}, the model faces domain shifts across datasets: Compared with the performance on the Masa dataset, the performance of both methods decreases $0.0375$ on average. However, GeoExplorer still outperforms the baseline, suggesting a better cross-domain transferability.

\begin{table}[t]
    \centering
    \small
    \caption{\textbf{Cross-domain transfer evaluation} on the \textbf{SwissView100 subset of SwissView dataset}.}
    \resizebox{0.46\textwidth}{!}{
    \begin{tabular}{lccccc}
        \toprule
        Method & $\mathcal{C}=4$ & $\mathcal{C}=5$ & $\mathcal{C}=6$ & $\mathcal{C}=7$ & $\mathcal{C}=8$  \\
        \cmidrule(r){1-6}
        GOMAA-Geo$^*$ & \textbf{0.4100} & 0.5000 & 0.6580 & 0.7780 & 0.6880\\
        \textbf{GeoExplorer} & 0.4020 & \textbf{0.5120} & \textbf{0.7660} & \textbf{0.9040} & \textbf{0.8800}\\
        \bottomrule
    \end{tabular}}
    \label{tab:swissview}
\end{table}

\begin{table}[t]
    \centering
    \small
    \caption{\textbf{Supplementary results} on the \textbf{MM-GAG dataset} with 65 search areas. The goal is presented as an aerial image (``I''), a ground-level image (``G''), or a text (``T''). Results from the original paper are in gray and $^*$ denotes the results on the same configurations using pretrained model provided by the paper.}
    \resizebox{0.48\textwidth}{!}{
    \begin{tabular}{clccccc}
        \toprule
        Goal & Method & $\mathcal{C}=4$ & $\mathcal{C}=5$ & $\mathcal{C}=6$ & $\mathcal{C}=7$ & $\mathcal{C}=8$  \\
        \cmidrule(r){1-7}
          \multirow{3}{*}{I}   & \textcolor{gray}{GOMAA-Geo~\cite{sarkar2024gomaa}} & \textcolor{gray}{0.4085} & \textcolor{gray}{0.5064} & \textcolor{gray}{0.6638} & \textcolor{gray}{0.7362} & \textcolor{gray}{0.7021}\\
        & GOMAA-Geo$^*$ & 0.4246 & 0.4769 & 0.7385 & 0.7662 & 0.6369\\
        & \textbf{GeoExplorer} & \textbf{0.4338} & \textbf{0.5415} & \textbf{0.7631} & \textbf{0.8369} & \textbf{0.8277}\\ 
        \multirow{3}{*}{G} & \cellcolor{mygrey}\textcolor{gray}{GOMAA-Geo~\cite{sarkar2024gomaa}} & \cellcolor{mygrey}\textcolor{gray}{0.4383} & \cellcolor{mygrey}\textcolor{gray}{0.5150} & \cellcolor{mygrey}\textcolor{gray}{0.6808} & \cellcolor{mygrey}\textcolor{gray}{0.7489} & \cellcolor{mygrey}\textcolor{gray}{0.6893}  \\ 
        & \cellcolor{mygrey}GOMAA-Geo$^*$ & \cellcolor{mygrey}\textbf{0.4585} & \cellcolor{mygrey}0.4554 & \cellcolor{mygrey}0.6646 & \cellcolor{mygrey}0.7169 & \cellcolor{mygrey}0.6708\\
        & \cellcolor{mygrey}\textbf{GeoExplorer} & \cellcolor{mygrey}0.4308 & \cellcolor{mygrey}\textbf{0.5138} & \cellcolor{mygrey}\textbf{0.7200} & \cellcolor{mygrey}\textbf{0.8246} & \cellcolor{mygrey}\textbf{0.7815}\\ 
        \multirow{3}{*}{T} & \textcolor{gray}{GOMAA-Geo~\cite{sarkar2024gomaa}} & \textcolor{gray}{0.4000} & \textcolor{gray}{0.4978} & \textcolor{gray}{0.6766} & \textcolor{gray}{0.7702} & \textcolor{gray}{0.6595}\\
        & GOMAA-Geo$^*$ & 0.4277 & \textbf{0.5015} & 0.6523 & 0.7538 & 0.6677\\
        & \textbf{GeoExplorer} & \textbf{0.4431} & 0.4892 & \textbf{0.7200} & \textbf{0.8062} & \textbf{0.7631}\\
        \bottomrule
    \end{tabular}}
    \label{tab: suppmmgag}
\end{table}

\subsection{Supplementary results on the MM-GAG dataset}
As mentioned in Section~\ref{ssec:dataset}, only 65 search areas are found through the link provided by the paper. To ensure fair comparison, we evaluate GeoExplorer and the pre-trained GOMAA-Geo on the same test configurations and report the results in Table~\ref{tab: suppmmgag}. We also report the GOMAA-Geo performance from the original paper for reference. The results suggest similar observations with 65 and 73 search areas: GeoExplorer achieves performances comparable to the baseline on short paths, while significantly improving SR when the path is longer.

\begin{figure}[t]
    \centering
\includegraphics[width=0.9\linewidth]{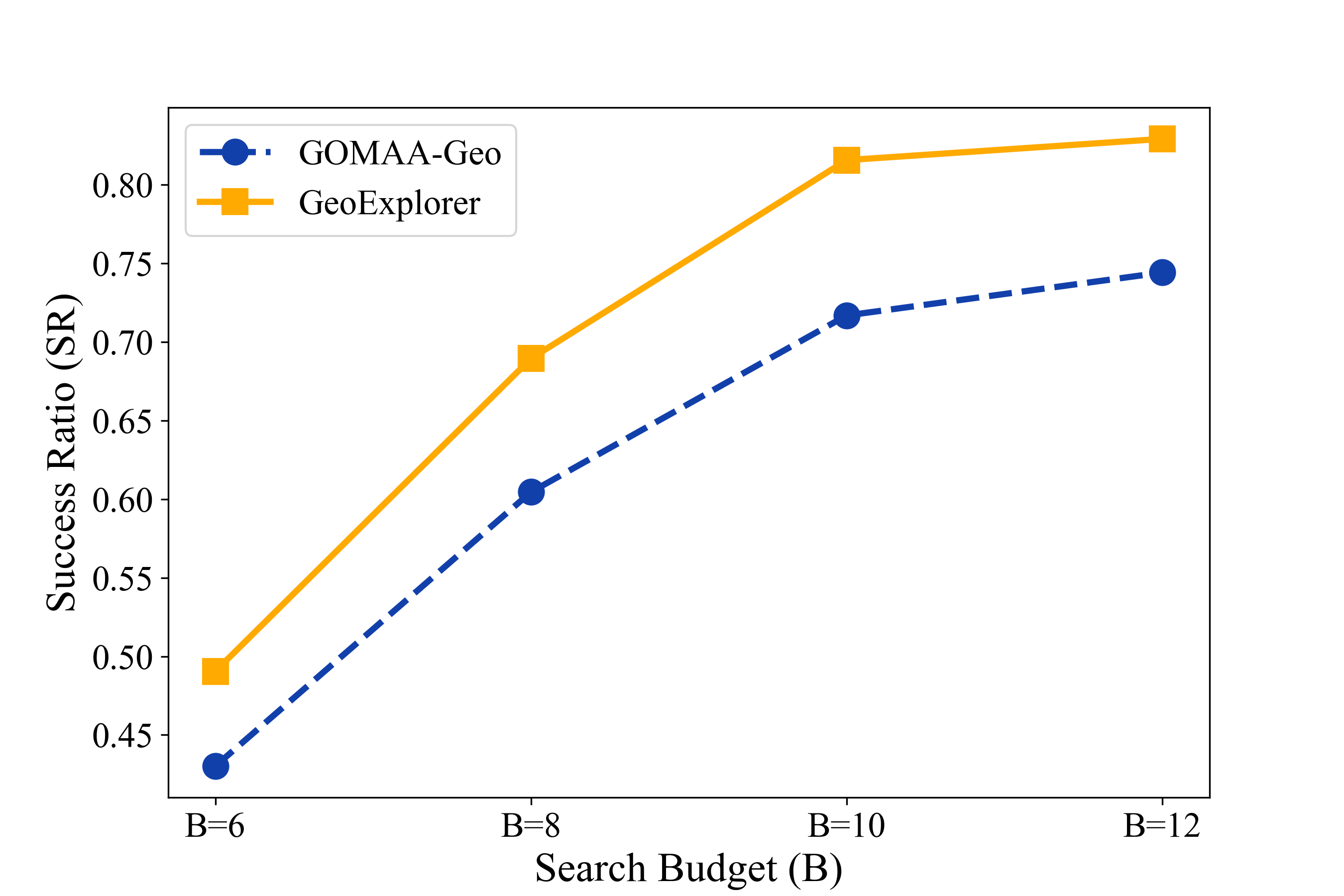}
    \caption{Comparison between GeoExplorer and the baseline model with \textbf{varying search budget} when $\mathcal{C}=6$.} 
    \label{fig:budget}
\end{figure}

\subsection{Exploration with varying search budget}
Search budget ($\mathcal{B}$) is an important factor in AGL. To give further insights on its impact on the exploration behaviour, we compare GeoExplorer and the baseline model GOMAA-Geo with varying $\mathcal{B}$, when $\mathcal{C}=6$ in Figure~\ref{fig:budget}. As expected, as $\mathcal{B}$ increases, the performance of both models improves, as the tolerance for mistakes is higher. More interestingly, the advantage of GeoExplorer is more obvious when $\mathcal{B}$ increases ($0.0603$ when $\mathcal{B}=6$ while $0.0850$ when $\mathcal{B}=12$), as it allows more exploration during the localization process.

\subsection{Exploration with larger grid size}
We compare GeoExplorer and GOMAA-Geo~\cite{sarkar2024gomaa} on a grid size of $10\times10$ with $\mathcal{C}=\{14-18\}$, providing \textit{more variability for longer paths}. As shown in Table~\ref{tab: grid}, GeoExplorer consistently shows improvements, especially for longer paths, which aligns with results tested on $5\times5$. Note that all the models are retrained for the grid size of $10\times10$.

\begin{table}[t]
    \centering
    \caption{Comparison between GeoExplorer and the baseline model with \textbf{larger grid size} on the Masa dataset. Note that all the methods are trained on the $10\times10$ grid size. $^{\star\star}$ corresponds to results obtained from the re-trained model using official code~\cite{sarkar2024gomaa}.}
    \small
     \resizebox{0.95\columnwidth}{!}{
    \begin{tabular}{lccccc}
        \toprule
        Method& $\mathcal{C}=14$ & $\mathcal{C}=15$ & $\mathcal{C}=16$ & $\mathcal{C}=17$ & $\mathcal{C}=18$ \\
        \cmidrule(r){1-6} 
        GOMAA-Geo$^{\star\star}$ & 0.2603 & 0.2704 & 0.2916 & 0.2413 & 0.2201 \\ 
        \textbf{GeoExplorer} & \textbf{0.2883} & \textbf{0.3117} & \textbf{0.3352} & \textbf{0.3073} & \textbf{0.3151}\\ 
        \bottomrule
    \end{tabular}
    }
    \label{tab: grid}
\end{table}

\subsection{Step-to-the-goal (SG) Evaluation}
Besides the commonly used metric success ratio (SR) in AGL, we also provide step-to-the-goal (SG) to evaluate the Manhattan distance between the path-end and goal locations on the SwissView dataset. Results in the Table~\ref{tab: sg} indicate that GeoExplorer \textit{improves success rate and brings the agent much closer to the goal}.

\begin{table}[t]
    \centering
    \caption{\textbf{Step-to-the-goal evaluation of unseen objects generalization ability} on the \textbf{SwissViewMonuments dataset}. $^{\star}$ corresponds to results obtained from the pretrained model ~\cite{sarkar2024gomaa}.}
    \small
     \resizebox{0.96\columnwidth}{!}{
    \begin{tabular}{clccccc}
        \toprule
        &  Method & $\mathcal{C}=4$ & $\mathcal{C}=5$ & $\mathcal{C}=6$ & $\mathcal{C}=7$ & $\mathcal{C}=8$ \\
        \cmidrule(r){1-7} 
        \multirow{2}{*}{I} & GOMAA-Geo$^{\star}$  & 2.16 & 1.91 & 1.18 & 0.93 & 0.77 \\ 
        & \textbf{GeoExplorer} & \textbf{2.08} & \textbf{1.56} & \textbf{0.65} &  \textbf{0.28} & \textbf{0.30}\\ 
        \multirow{2}{*}{G} &  \cellcolor{mygrey}GOMAA-Geo$^{\star}$  & \cellcolor{mygrey}2.19 &  \cellcolor{mygrey}1.81 & \cellcolor{mygrey}1.07 &  \cellcolor{mygrey}0.69 &  \cellcolor{mygrey}0.77 \\ 
        & \cellcolor{mygrey}\textbf{GeoExplorer} & \cellcolor{mygrey}\textbf{2.14} &  \cellcolor{mygrey}\textbf{1.51} & \cellcolor{mygrey}\textbf{0.64}  & \cellcolor{mygrey}\textbf{0.29} & \cellcolor{mygrey}\textbf{0.70}\\
        \bottomrule
    \end{tabular}
    }
    \label{tab: sg}
\end{table}

\section{Suppelmentary Analysis}
\label{sec:suppana}
\subsection{Path statistics}
\begin{figure*}[t]
    \centering
    \includegraphics[width=0.96\linewidth]{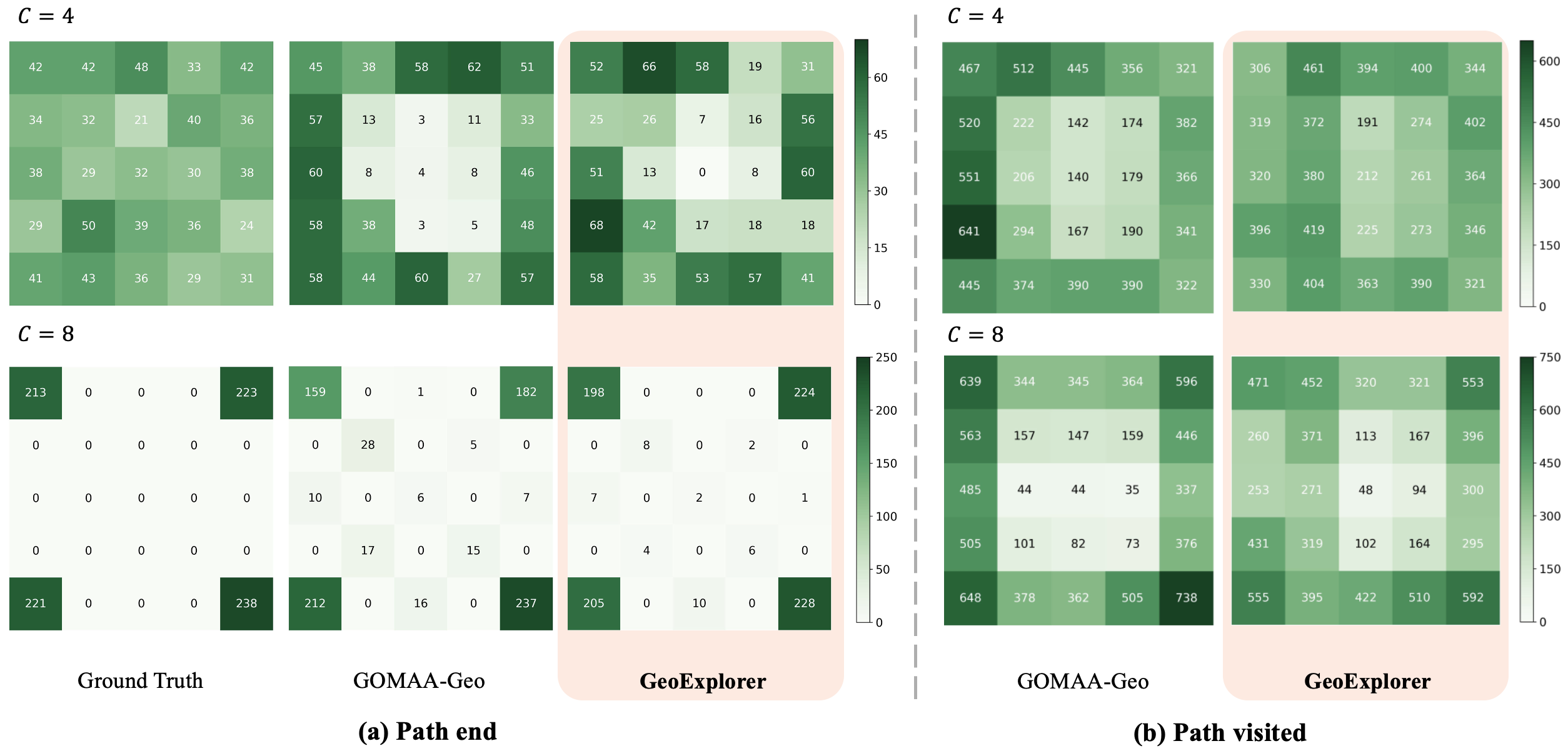}
    \caption{\textbf{GeoExplorer tends to explore more patches in the search areas, especially the inside patches.} (a) \textit{Statistics of the path end.} We count the end location of the 895 paths in the Masa dataset test set for ground truth (goal location), GOMAA-Geo and GeoExplorer when $\mathcal{C}=4$ and $\mathcal{C}=8$. (b) \textit{Statistics of the path visited.} We count all the visited patches of 895 paths in the Masa dataset test set for GOMAA-Geo and GeoExplorer when $\mathcal{C}=4$ and $\mathcal{C}=8$.} 
    \label{fig:path_statisics}
\end{figure*}

We provide an in-depth analysis of the exploration ability of the model by tracking the visited patches of the baseline model and GeoExplorer. In Figure~\ref{fig:path_statisics} (a), we count the end location of 895 paths from the Masa test set for the ground truth (goal location), GOMAA-Geo and GeoExplorer. The results confirm that 1) for $\mathcal{C}=4$ the goal locations are more evenly distributed in the search area while the configurations are limited for $\mathcal{C}=8$. This could explain why we usually have higher performance when $\mathcal{C}=8$. As the trajectory grows, the agent may infer that the goal is likely located at the corner, based on the training data distribution. 2) The path end distribution of the baseline model suggests a tendency to visit edge patches (with $84.36\%$ of paths ended on the edge patches when $\mathcal{C}=4$), while GeoExplorer improves the exploration of inside patches. The visited patch distribution shown in Figure~\ref{fig:path_statisics} (b) confirms this observation: when $\mathcal{C}=4$, only $20.08\%$ of the visited patches are inside for GOMAA-Geo and GeoExplorer increases this ratio as $30.79\%$. This finding indicates that Geoexplorer improves the performance on the AGL benchmarks by a better exploration ability of the environment.

\begin{figure}[t]
    \centering
    \includegraphics[width=0.9\linewidth]{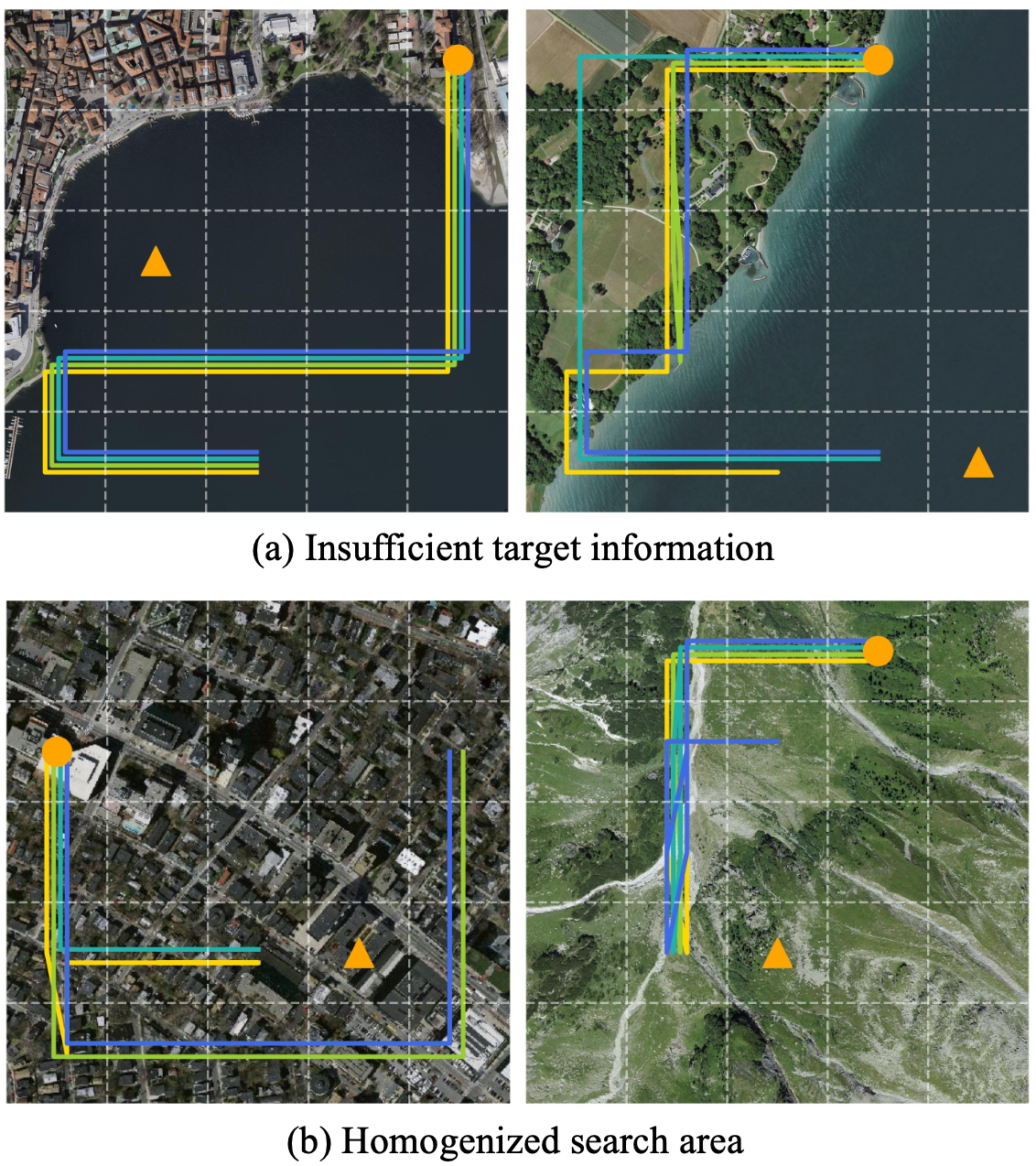}
    \caption{\textbf{Failure case analysis.} (a) Insufficient target information. The target provides limited information for the localization. (b) Homogenized search area. Patches in the search area are similar and could mislead the exploration.} 
    \label{fig:failure}
\end{figure}

\subsection{Failure case analysis}
We check and analyze the failure cases of GeoExplorer. Examples shown in Figure~\ref{fig:failure} imply two main reasons of failure: (a) Insufficient target information and (b) Homogenized search area. In Figure~\ref{fig:failure} (a), the goals are located in the water and are very similar to the surrounding patches, providing limited localization information. Along the generated path, several similar patches have been visited, but not the target patch. In Figure~\ref{fig:failure} (b), all patches are similar in the search area (e.g., similar building patches in an urban area or mountain patches), which could probably mislead the exploration process.

\subsection{Additional path visualization examples} 

\begin{figure*}[t]
    \centering
    \includegraphics[width=0.95\linewidth]{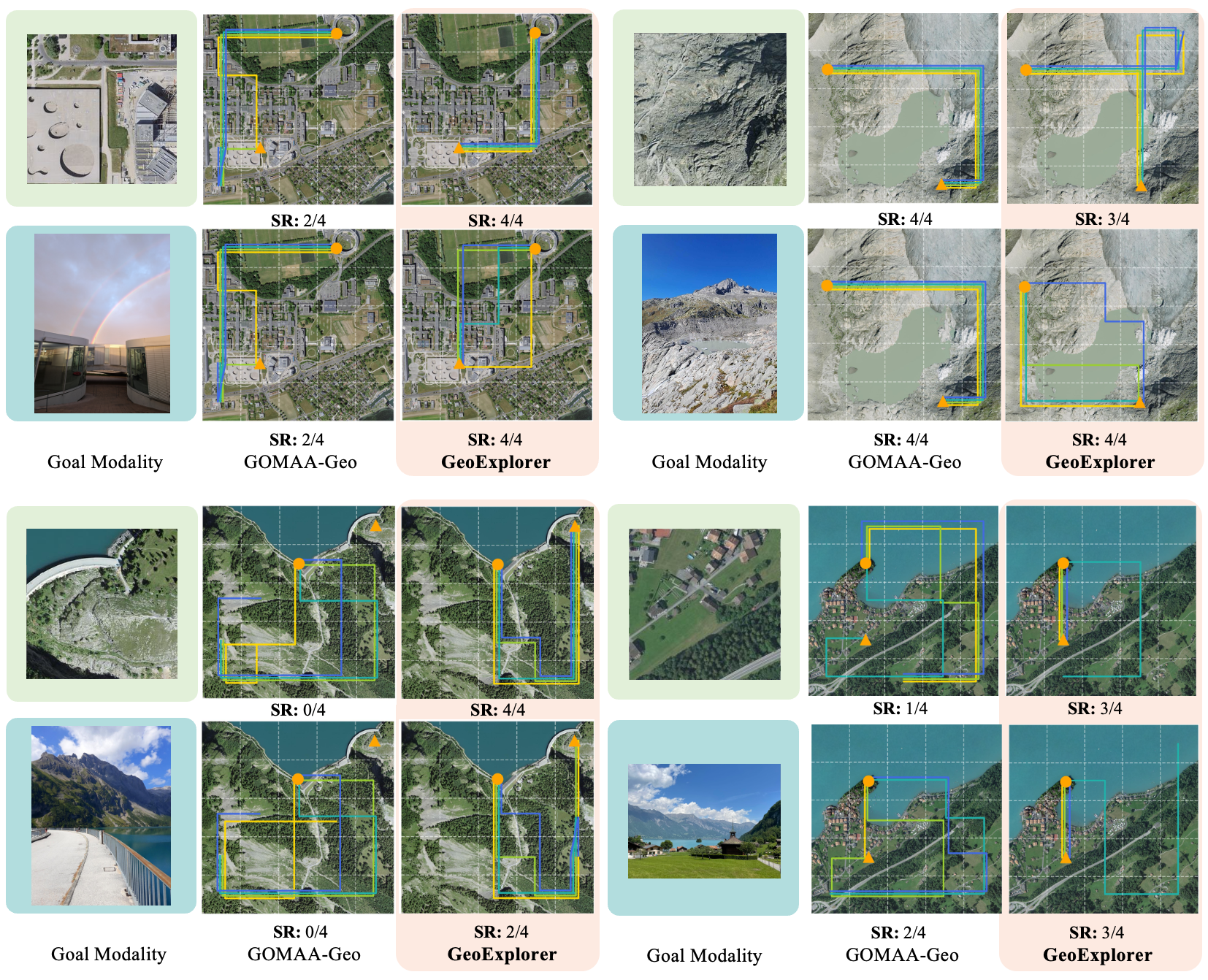}
    \caption{\textbf{Path visualization from the SwissViewMonuments dataset} with goals presented in aerial and ground views.} 
    \label{fig:path_unseen}
\end{figure*}

\begin{figure*}[t]
    \centering
    \includegraphics[width=0.95\linewidth]{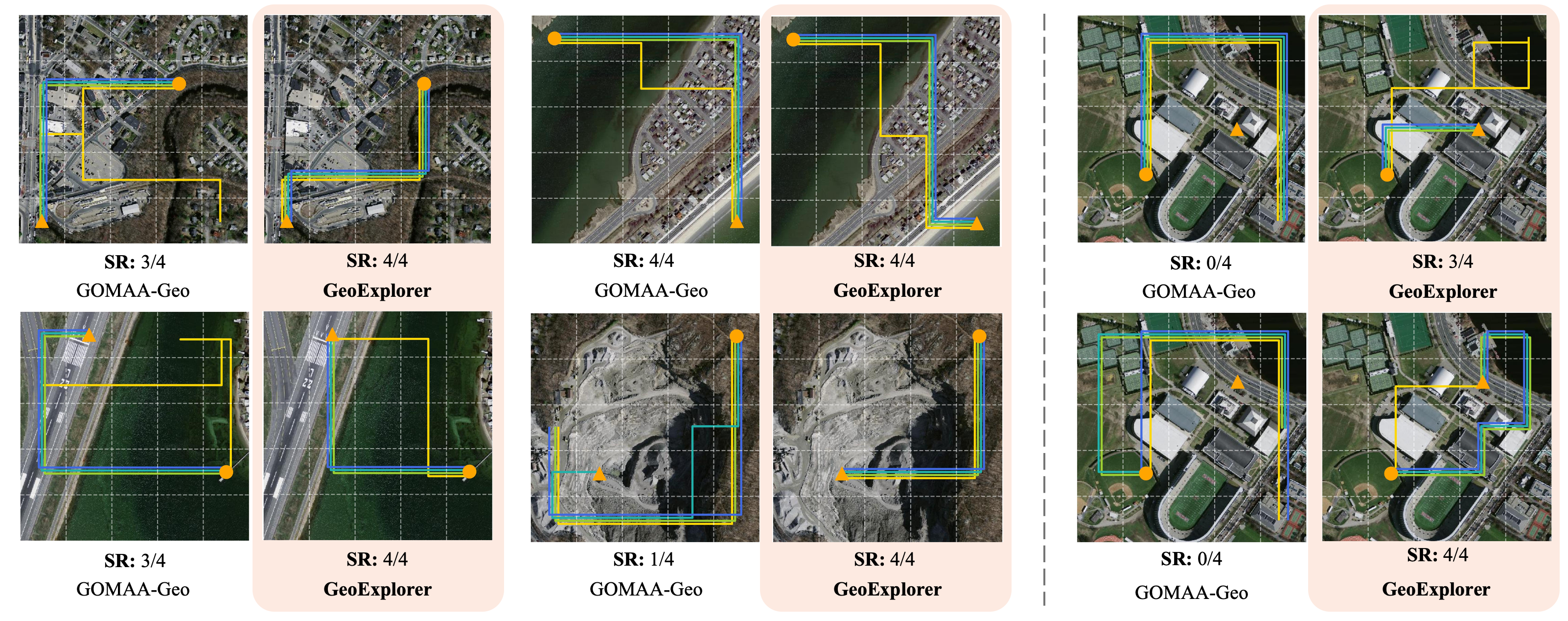}
    \caption{\textbf{Path visualization from the Masa dataset test set} with goals presented in aerial view. On the right, we show examples with more controllable $\{$start, goal$\}$ configuration: the location of the goal patches changed slightly and the start patch remains the same.} 
    \label{fig:path_masa}
\end{figure*}

\begin{figure*}[t]
    \centering
    \includegraphics[width=0.95\linewidth]{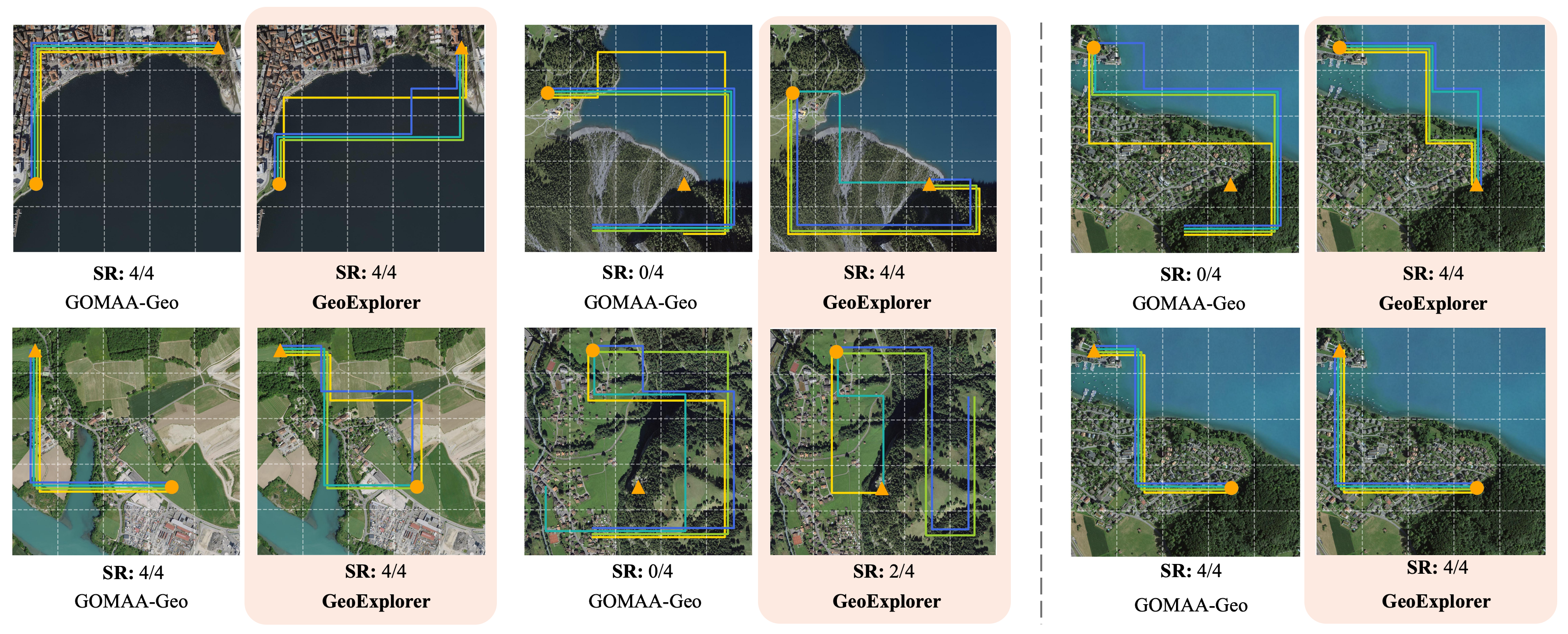}
    \caption{\textbf{Path visualization from the SwissView100 dataset} with goals presented in aerial view. On the right, we show examples with more controllable $\{$start, goal$\}$ configurations: with reversed start and goal locations.} 
    \label{fig:path_swissview}
\end{figure*}

To further support our findings, we provide additional visualization examples from the SwissViewMonuments dataset (Figure~\ref{fig:path_unseen}), the Masa dataset (Figure~\ref{fig:path_masa}) and the SwissView100 dataset (Figure~\ref{fig:path_swissview}). These figures illustrate key aspects of GeoExplore's performance, in particular in terms of adaptation to new environments and targets and better exploration strategies, and confirm the quantitative results presented before. In particular, the three figures show that \textbf{(1)} GeoExplorer achieves higher success rates (SR) and generalizes better in novel environments (Figure~\ref{fig:path_swissview}, SwissView100 examples) and when faced with unusual and unseen goals (Figure~\ref{fig:path_unseen}, SwissViewMonuments examples). \textbf{(2)} GeoExplorer produces also more diverse paths, while the baseline model (GOMAA-Geo) tends to follow edge patches, often navigating towards corners before heading to the goal. This aligns with the statistical observations made in the main paper, which highlighted that the goals were more frequently located on the edges and corners and may lead GOMAA-Geo to overfit in these areas. \textbf{(3)} Visualization samples also indicate the robustness of GeoExplorer's exploration. The right panels of Figures~\ref{fig:path_masa} illustrate how a slight change in the goal location affects the exploration process for both the Masa dataset. GeoExplorer seems to be more robust and adapt its exploration, demonstrating diverse and flexible path selection. In contrast, the baseline model tends to follow similar paths regardless of these small changes. The right panels of Figure~\ref{fig:path_swissview} show another controllable configuration on the SwissView100 dataset, where we reverse start and goal locations between the upper and lower examples, which demonstrates GeoExplorer's exploration is more robust to this reversion. Moreover, the images in the two figures are sourced from different platforms, have varying resolutions, and depict different locations. This highlights how the proposed dataset enhances data diversity for the task.

\subsection{Additional intrinsic reward visualization examples}
\begin{figure*}[t]
    \centering
    \includegraphics[width=0.95\linewidth]{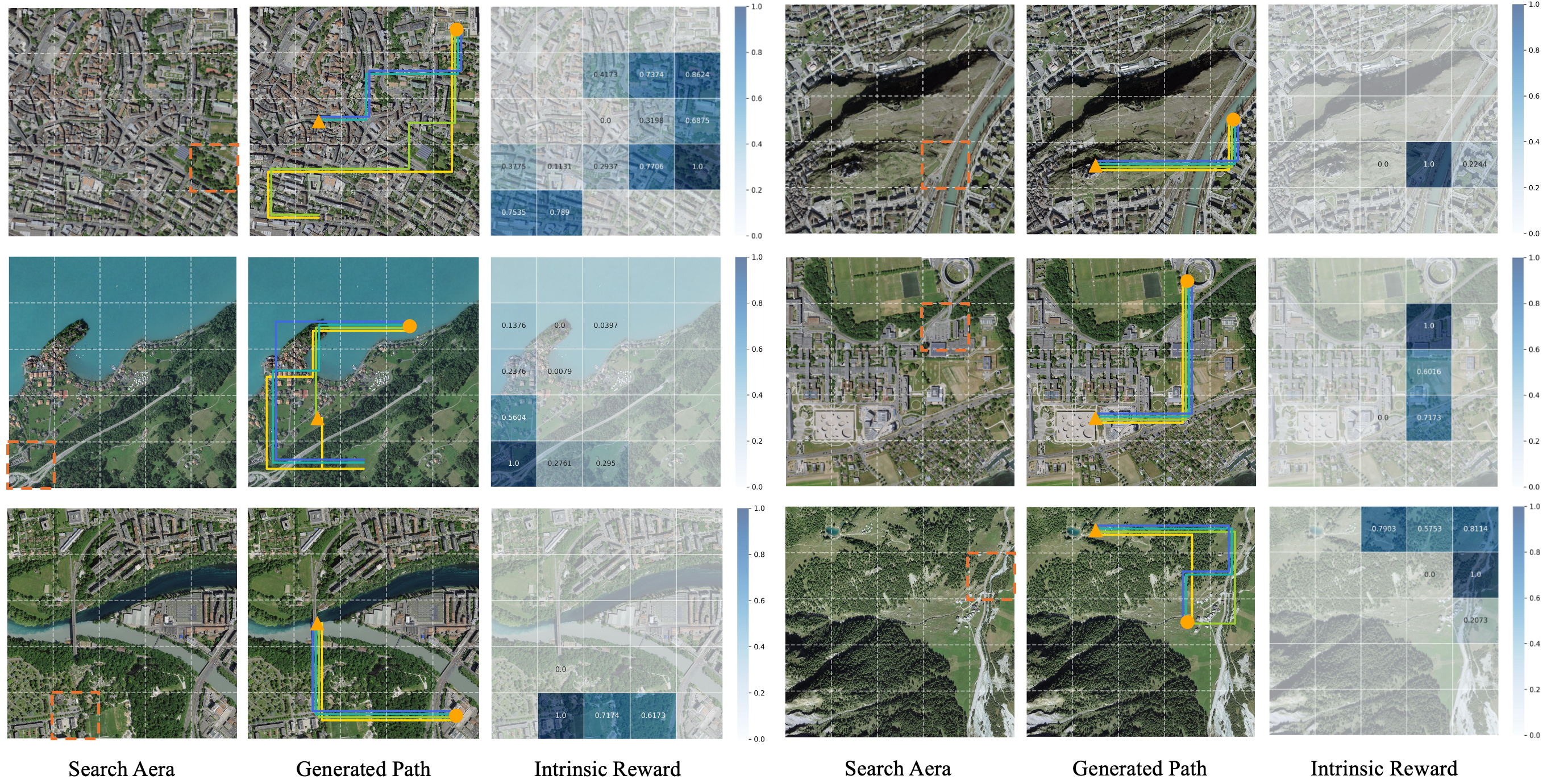}
    \caption{\textbf{Intrinsic reward visualization with images from the SwissViewMonuments dataset}. For each sample, from left to right: the search area, path visualization and intrinsic reward per patch. The patch with the highest intrinsic reward is highlighted with an orange rectangle in the search area.} 
    \label{fig:suppreward}
\end{figure*}

To provide further insights to the intrinsic reward, we provide additional samples from the SwissViewMonuments dataset in Figure~\ref{fig:suppreward}. The patches with higher intrinsic rewards are usually unique patches in the search area (\textit{e.g.}, the first example of the right column: a green land in an urban region) or the surprising sample along the path (e.g., the first example of the left column: moving from an urban patch to river). The findings indicate the intrinsic rewards are content-aware and improve the model's exploration ability with dense and goal-agnostic guidance.

\section{Discussion}
\label{sec:disscusion}
As an emerging research topic, the task configuration of AGL still includes some limitations: (1) Continuous state and action space. Currently, AGL considers a grid-like environment space for states and actions. For example, different states have no overlaps and actions are chosen from a discrete space. This setting could be further improved as a continuous space for both states and actions to meet the requirements of real-world search-and-rescue operations.
(2) Real-world development. When developing the models on an UAV agent, there are some other challenges, for example, the noisy ego-pose of the agent and the observation deformation. Those challenges should also be considered for further real-world development of AGL tasks.

As for the methodology, the proposed Curiosity-Driven Exploration has shown impressive exploration ability for the AGL task, and would encourage future work to further understand the exploration pattern of an agent. For example, an in-depth study and analysis of how intrinsic reward affects extrinsic reward and a comprehensive analysis of how to combine those two motivations would further provide insights to not only AGL but also other goal-reaching reinforcement learning tasks.

\clearpage

\clearpage
\clearpage
\newpage

{
    \small
    \bibliographystyle{ieeenat_fullname}
    \bibliography{main}
}

\end{document}